\documentclass[10pt,twocolumn,letterpaper]{article}

\usepackage{cvpr}              
\definecolor{cvprblue}{rgb}{0.21,0.49,0.74}
\usepackage[pagebackref,breaklinks,colorlinks,allcolors=cvprblue]{hyperref}

\usepackage{multirow}
\usepackage{makecell}
\usepackage{bm}
\usepackage{colortbl}
\usepackage{color,xcolor}
\usepackage{pifont}
\usepackage{float}

\def\paperID{***} 
\def\confName{CVPR}
\def\confYear{2026}

\title{Sparse Task Vector Mixup with Hypernetworks for Efficient Knowledge Transfer\\ in Whole-Slide Image Prognosis}

\author{Pei Liu$^{1}$, Xiangxiang Zeng$^{1,}$\thanks{Correspondence to X. Zeng ({\tt xzeng@hnu.edu.cn})} , Tengfei Ma$^{1}$, Yucheng Xing$^{2}$, 
Xuanbai Ren$^{1}$, Yiping Liu$^{1}$\\
{\small ${^1}$College of Computer Science and Electronic Engineering, Hunan University~~~${^2}$National University of Singapore}
}

\newcommand{\ours}{\textsc{STEPH}}

\begin{document}
\maketitle
\begin{abstract}

Whole-Slide Images (WSIs) are widely used for estimating the prognosis of cancer patients.
Current studies generally follow a cancer-specific learning paradigm.
However, the available training samples for one cancer type are usually scarce in pathology.
Consequently, the model often struggles to learn generalizable knowledge, thus performing worse on the tumor samples with inherent high heterogeneity.
Although multi-cancer joint learning and knowledge transfer approaches have been explored recently to address it, they either rely on large-scale joint training or extensive inference across multiple models, posing new challenges in computational efficiency.
To this end, this paper proposes a new scheme, \underline{S}parse \underline{T}ask V\underline{e}ctor Mixu\underline{p} with \underline{H}ypernetworks ($\ours$). Unlike previous ones, it efficiently absorbs generalizable knowledge from other cancers for the target via model merging: i) applying task vector mixup to each source-target pair and then ii) sparsely aggregating task vector mixtures to obtain an improved target model, driven by hypernetworks.
Extensive experiments on 13 cancer datasets show that $\ours$ improves over cancer-specific learning and an existing knowledge transfer baseline by 5.14\% and 2.01\%, respectively.
Moreover, it is a more efficient solution for learning prognostic knowledge from other cancers, without requiring large-scale joint training or extensive multi-model inference.
Code is publicly available at \url{https://github.com/liupei101/STEPH}.

\end{abstract}
\section{Introduction}
\label{sec:intro}

Histopathology Whole-Slide Image (WSI) is a special medical image with gigapixel size (\eg, 20,000 $\times$ 20,000).
It can exhibit microscopic details in tumor tissues at different levels, \eg, cellular morphology, cell-tissue interactions, and tumor micro-environments~\cite{chen2022scaling,song2024morphological}.
These comprehensive features enable doctors to scientifically assess the progression of tumor diseases and estimate the prognosis of cancer patients~\cite{yu2016predicting,Kather2019} (also called \emph{survival analysis}).
Since a precise prognosis could provide critical guidance for therapeutic planning~\cite{chen2025whole}, various computational approaches have been developed to model WSIs for accurate prognosis estimates~\cite{chen2021whole,shao2021weakly,liu2024advmil,pmlr-v235-song24b,liu2025interpretable,xu2025distilled,zhou2025robust,wu2025learning}.

\begin{figure}[t]
  \centering
   \includegraphics[width=\linewidth]{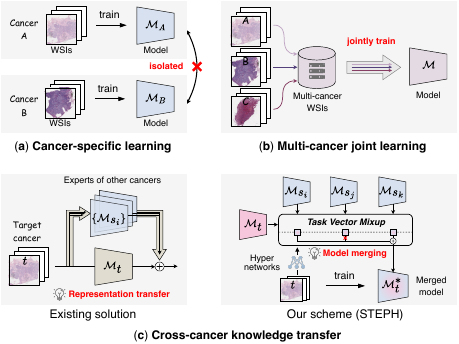}
   \caption{\textbf{Current learning paradigms of modeling WSIs for survival analysis}. Unlike previous approaches, the proposed $\ours$ efficiently utilizes the generalizable prognostic knowledge from other cancers by model merging.}
   \label{fig:1_existing_frameworks}
\end{figure}

Cancer-specific learning, as depicted in \cref{fig:1_existing_frameworks} (a), has always been the mainstream paradigm~\cite{lu2023visual} for developing prognostic models.
In this line of research, current studies focus on devising better multi-instance learning (MIL) approaches~\cite{ilse2018attn,li2021dual,zhang2022dtfd,shao2021transmil} to gather key visual information from ultra-large WSIs.
They have demonstrated continuously improving abilities in capturing prognosis-relevant cues.
However, WSI samples are often scarce for one cancer type ($N\approx 1,000$) in the real world.
This naturally raises a common concern: could the prognostic model trained on very limited data \emph{generalize well}, especially when confronted with high heterogeneity in tumors?

Multi-cancer joint learning~\cite{yuan2025pancancer,zhou2025multimodal} is a straightforward solution for the above issue.
As shown in \cref{fig:1_existing_frameworks} (b), instead of training a cancer-specific model on limited samples, it constructs a multi-cancer WSI dataset to scale up the data size and \emph{jointly} train a model that can learn generalizable features across different tumors.
Nevertheless, this approach is rooted in large-scale multi-cancer data.
Due to the gigapixel size of WSIs, it not only incurs extremely expensive computational costs, but also raises potential concerns about data privacy~\cite{wei2024task}.
These intractable issues call for a new learning paradigm that enables the model to utilize the knowledge from other cancers more efficiently.

\begin{figure}[t]
  \centering
   \includegraphics[width=\linewidth]{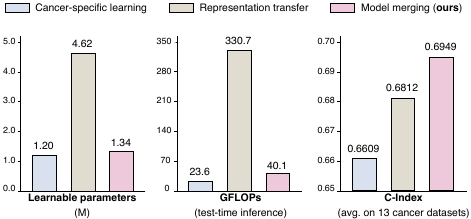}
   \caption{\textbf{Performance of different learning paradigms} in WSI prognosis. We compare $\ours$ with cancer-specific learning and a representation transfer-based solution. Multi-cancer joint learning is absent as its training overheads and hardware specifications are orders of magnitude greater than those of others.}
   \label{fig:2_perf_overview}
\end{figure}

Cross-cancer knowledge transfer, \ie, ROUPKT~\cite{liu2025cross}, has been proposed recently to fulfill the above purpose.
Instead of directly training on a gigantic multi-cancer WSI dataset, it employs multiple off-the-shelf models and transfers their generalizable knowledge for a target cancer by integrating the image \emph{representations} output by them, as shown in \cref{fig:1_existing_frameworks} (c).
Although the potential of knowledge transfer has been proven, such a representation-based solution still faces critical challenges in terms of computation efficiency: it has to explicitly pass each sample through multiple transferred models at test-time, causing the computational overhead to increase linearly with the number of transferred models.
This limitation motivates us to ask: \textit{can we ultimately use one single model to fulfill efficient cross-cancer knowledge transfer}?

To answer this question, we take inspiration from \emph{model merging}~\cite{ilharco2023editing}, a lightweight approach for merging multiple source models into one target model that can reuse or transfer knowledge from different sources.
Built upon its fundamentals, \ie, task vectors $\tau_t=\mathcal{M}_t-\mathcal{M}_{\text{0}}$, we propose a new scheme, \underline{S}parse \underline{T}ask V\underline{e}ctor Mixu\underline{p} with \underline{H}ypernetworks (\textbf{$\ours$}), for efficient cross-cancer knowledge transfer.
As illustrated in \cref{fig:1_existing_frameworks} (c), it merges the model of a target cancer ($\mathcal{M}_t$) and those of other cancers ($\{\mathcal{M}_{s_i}\}$) into a single $\mathcal{M}_t^{*}$ by \emph{i}) applying task vector mixup to each paired $\tau_t$ and $\tau_{s_i}$ and then \emph{ii}) sparsely aggregating task vector mixtures to obtain $\mathcal{M}_t^{*}$. 
Hypernetworks are devised to drive this process. They learn how to adjust the weight of $\mathcal{M}_t^{*}$ by steering task vectors.
In this way, $\mathcal{M}_t^{*}$ can absorb the knowledge from other cancers, thereby improving the generalizability of $\mathcal{M}_t$.
Unlike previous approaches, $\ours$ leverages cross-cancer knowledge more efficiently, without requiring either large-scale joint training or intensive multi-model inference.
We compare $\ours$ with different learning paradigms using 13 datasets (see \cref{fig:2_perf_overview}). Results show that $\ours$ outperforms traditional cancer-specific learning and a representation transfer-based solution by 5.14\% and 2.01\%, respectively.
It only introduces marginal additional costs and has significantly lower computation overhead than the existing solution.

Our contributions are summarized as follows.
(1) We propose Sparse Task Vector Mixup with Hypernetworks ($\ours$) to fulfill efficient knowledge transfer across cancers in WSI prognosis.
(2) Task vector mixup (TVM) is presented as a variant of task arithmetic tailored for transferring knowledge from one cancer to another; a principled analysis and empirical studies suggest that TVM enhances the generalizability of prognostic models by offering better optimization directions.
(3) We conduct extensive comparative experiments on 13 cancer datasets; results show that $\ours$ often efficiently transfers knowledge from other cancers and performs better than existing methods.

\section{Related work}
\label{sec:related_work}

\subsection{WSI-based survival analysis}

Due to the gigapixel size, histopathology WSIs are usually divided into image patches, followed by patch feature extraction with pretrained foundation models~\cite{chen2024towards,lu2024visual,xiang2025vision}. This produces a bag of numerous feature vectors (or instances) for one WSI.
To learn a slide-level representation from it for survival analysis, various multi-instance learning (MIL) approaches with the cluster~\cite{yao2020whole,xing2025dpsurv}, graph~\cite{chen2021whole,shao2024tumor}, or pure sequence~\cite{ilse2018attn,shao2021transmil,liu2025interpretable} structure imposed on instances, are proposed to aggregate global prognosis-relevant features from local instances under the supervision of time-to-event labels.
Despite their remarkable success, these studies generally follow a conventional cancer-specific learning paradigm: one model corresponds to one specific cancer, and this model does not benefit from the generalizable knowledge of other cancers.
To tackle it, multi-cancer joint learning~\cite{yuan2025pancancer,zhou2025multimodal} and cross-cancer knowledge transfer~\cite{liu2025cross} have been studied recently.
However, these solutions still struggle to learn generalizable prognostic knowledge from other cancers efficiently.

\subsection{Model merging}

Model merging~\cite{wortsman2022robust,wortsman2022model,ilharco2023editing} emerges as an efficient approach that reuses or transfers knowledge from models by editing the weights of neural networks.
\citet{ilharco2023editing} propose \emph{task vectors} $\tau_t$, obtained by subtracting the weights of a pretrained model from the weights of the same model after fine-tuning on task $t$, to represent the knowledge learned from $t$.
Since its inception, various task vector-based methods have been investigated for certain purposes; multi-task learning (MTL) attracts the most attention~\cite{tang2024fusionbench}.
Instead of performing MTL on large-scale datasets, the task vectors of task-specific models are computed and aggregated to obtain a multi-task model without training.
However, simply aggregating task vectors often results in limited multi-task performances due to the underlying \emph{interference} between tasks, so a multitude of methods are proposed to resolve task interference by disentangling task-common and task-specific subspaces~\cite{yadav2023ties,ortiz2023task,yangadamerging2024,yang2024representation,wang2024localizing,marczakno2025isoc,chen2025fwmergingscalingmodelmerging,wei2025modeling}.
Although this paper also studies model merging, it differs from the above studies in research purpose: it does \emph{not} aim to obtain a model capable of multiple tasks but aims to further enhance the generalizability of \emph{cancer-specific models} by model merging.
This discrepancy in purpose naturally leads to two distinct focuses on methodology: existing methods focus on addressing task interference, while our method focuses on transferring generalizable and beneficial knowledge from cross-cancer models for a target cancer.

\section{Preliminary}
\label{sec:preliminary}

This section introduces the preliminaries closely related to $\ours$, as well as some notations and conventions.

\subsection{General MIL architecture}

For a WSI, it is usually represented as a bag with multiple instances after patching and feature extraction, $X\in\mathbb{R}^{n\times d}$, where $x_i\in\mathbb{R}^d$ denotes an instance.

To learn a bag-level representation from instances, \citet{ilse2018attn} propose a general attention-based architecture.
It contains an instance embedding layer $f_{\text{emb}}(\cdot)$, followed by an attention layer $f_{\text{attn}}(\cdot)$. $f_{\text{emb}}(\cdot)$ encodes task-specific features for $x_i$ while $f_{\text{attn}}(\cdot)$ learns instance-wise attention scores and aggregates instance embeddings into a single vector as a bag-level representation by attention pooling.

\subsection{Task vectors}

\textbf{Definition}~Formally, given an initial model $\mathcal{M}_0$, $\mathcal{M}_t$ denotes a model obtained by training $\mathcal{M}_0$ on the data of task $t$.
The task vector of $\mathcal{M}_t$ is defined as $\tau_t=\mathcal{M}_t-\mathcal{M}_0$, which encodes the information necessary to do well on $t$~\cite{ilharco2023editing}.
From an optimization perspective~\cite{wei2025modeling}, $\tau_t$ represents cumulative \emph{gradients} for task $t$. It delineates an overall direction for a model to perform well on $t$.

\noindent\textbf{Task arithmetic}~Based on task vectors, a simple arithmetic operation is introduced in \cite{ilharco2023editing} to merge a set of individual task-specific models $\{\mathcal{M}_{t_i}\}$ into a new multi-task model:
\begin{equation}
    \mathcal{M}_{\text{new}}=\mathcal{M}_0+\tau_{\text{new}},
    \label{eqn:1}
\end{equation}
where $\tau_{\text{new}}=\sum_i\tau_{t_i}$.
Our model merging scheme is also built upon this fundamental operation, yet its focus is to produce a new model with better generalizability on $t$, rather than with multi-task capabilities.

\noindent\textbf{Task subspace alignment}~It is often of great interest to know whether a merged model $\mathcal{M}_{\text{new}}$ can perform well on a specific task $t$.
A common approach for this is to directly evaluate $\mathcal{M}_{\text{new}}$ using the test data of $t$.
As a more efficient alternative, Subspace Alignment Ratio (\textbf{SAR}) is studied in \cite{marczakno2025isoc}. It reflects how well $\mathcal{M}_{\text{new}}$ could perform on $t$ by quantifying the overlap between the dominant subspace of $\tau_{\text{new}}$ and $\tau_{t}$, as follows:
\begin{equation}
    \text{SAR}(\tau_{t},\tau_{\text{new}})=\frac{\|\text{Proj}_{\tau_{\text{new}}^{\alpha}}\tau_{t}\|_F}{\|\tau_{t}\|_F},
    \label{eqn:2}
\end{equation}
where $\tau_{\text{new}}^{\alpha}$ is the subspace spanned by the top $\alpha$ left-singular vectors of $\tau_{\text{new}}$, representing a dominant subspace of $\tau_{\text{new}}$ (\eg, $\alpha=0.95$).
$\text{Proj}_{\tau_{\text{new}}^{\alpha}}\tau_{t}$ is the projection of $\tau_{t}$ onto $\tau_{\text{new}}^{\alpha}$.
For the details of SAR calculation, please refer to Appendix.
It has been shown that $\text{SAR}(\tau_{t},\tau_{\text{new}})$ highly correlates with the performance of $\mathcal{M}_{\text{new}}$ on task $t$~\cite{marczakno2025isoc}.
In light of this, we adopt SAR to analyze the behavior and properties of new task vectors.

\section{Method}
\label{sec:method}

Given a target cancer type, written as $t$, we denote its corresponding prognostic model by $\mathcal{M}_t$.
Let $\{\mathcal{M}_{s_i}\mid i\in[1,m]\}$ represent a set of source models independently trained on cancer types $\{s_i\mid i\in[1,m]\}$.
To enhance the generalizability of $\mathcal{M}_t$, we propose $\ours$ (Sparse Task Vector Mixup with Hypernetworks) to efficiently transfer prognostic knowledge from other models $\{\mathcal{M}_{s_i}\mid i\in[1,m]\}$.

\begin{figure*}[t]
  \centering
   \includegraphics[width=\linewidth]{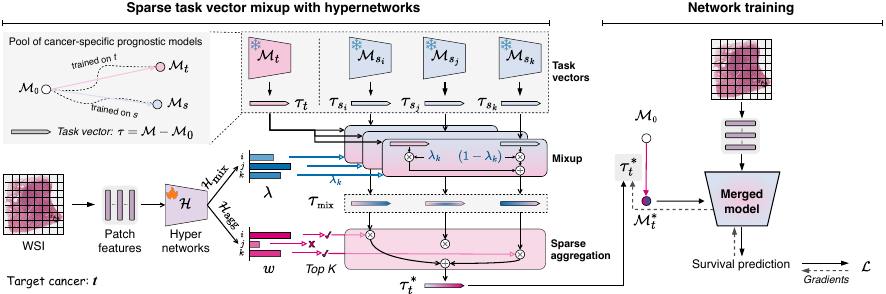}
   \caption{\textbf{Sparse Task Vector Mixup with Hypernetworks ($\ours$)} for efficient knowledge transfer in WSI prognosis. After computing task vectors, $\ours$ first applies mixup to each paired $(\tau_t,\tau_s)$ to absorb prognostic knowledge from cross-cancer models. Then, the most beneficial mixtures are selected and aggregated to derive $\mathcal{M}_t^{*}$ for prediction. Hypernetworks drive these steps by steering task vectors.}
   \label{fig:3_our_stvm}
\end{figure*}

As shown in \cref{fig:3_our_stvm}, $\ours$ is composed of three main steps: \textit{i}) task vector computation, \textit{ii}) task vector mixup, and \textit{iii}) sparse task vector aggregation. The last two are driven by hypernetworks. Next, we elucidate them, along with discussion and justification for key components.

\subsection{Task vector computation}

At first, task vectors are computed for all given models, as follows:
$\tau_t=\mathcal{M}_{t}-\mathcal{M}_{0}$ and $\tau_s=\mathcal{M}_{s}-\mathcal{M}_{0}$ for any $s\in\{s_i\mid i\in[1,m]\}$, according to the definition of task vectors~\cite{ilharco2023editing}, as shown in \cref{fig:3_our_stvm}.

\subsection{Task vector mixup (TVM)}

To absorb generalizable knowledge from $\{\mathcal{M}_{s_i}\}$ for $t$, our \emph{core idea} is Task Vector Mixup (TVM), \ie, applying mixup to any paired $\tau_{s_i}$ and $\tau_{t}$, inspired by the Vicinal Risk Minimization (VRM) principle~\cite{NIPS2000_ba9a56ce}.
Concretely, we \textit{i}) first construct a source-target model pair for any $s\in\{s_i\mid i\in[1,m]\}$ and \textit{ii}) then apply mixup interpolation to its task vectors, formulated as follows:
\begin{equation}
    \tau_{\text{mix}}=\lambda\tau_t+(1-\lambda)\tau_s.
    \label{eqn:3}
\end{equation}
$\lambda\in[0,1]$ is a mixup coefficient that determines the \emph{direction} of $\tau_{\text{mix}}$.
Intuitively, increasing $\lambda$ means pushing the direction of task vectors from $\tau_s$ to $\tau_t$.
Since $\tau_s$ encodes the knowledge of cancer $s$, \cref{eqn:3} could enable the target model to take in the knowledge from other models.

Next, we first \textit{i}) introduce our hypernetwork-driven TVM tailored for WSI prognosis and then \textit{ii}) justify TVM from a principled and an empirical perspective.

\subsubsection{Hypernetwork-driven TVM}

Although TVM could carry the generalizable knowledge from other cancers, determining the value of $\lambda$ remains a challenge.
A simple approach is to perform a grid search using a validation set. However, it tends to introduce large biases as the number of available WSI samples is typically around 1,000 for one specific cancer.
To tackle it, we employ a hypernetwork $\mathcal{H}_{\text{mix}}$ that learns to output adaptive $\lambda$ for different WSI inputs, rather than a fixed one for all inputs.
We call it ``\emph{hypernetwork}'' because it is a network that learns how to adjust the weights of another network (\ie, the final model) by steering task vectors.

As shown in \cref{fig:3_our_stvm}, for WSI patch features $X\in\mathbb{R}^{n\times d}$, an MIL-based hypernetwork $\mathcal{H}_{\text{mix}}$ is used to output $\lambda=\{\lambda_i\mid i\in[1,m]\}$, where $\lambda_i\in[0,1]$ corresponds to the adaptive mixup coefficient of a task vector pair $(\tau_t,\tau_{s_i})$.
Formally, we have $\tau_{\text{mix}}=\{\lambda_i\tau_t+(1-\lambda_i)\tau_{s_i}\mid i\in[1,m]\}$, where $\lambda_i\in\lambda=\mathcal{H}_{\text{mix}}(X)$.

\subsubsection{What is TVM doing?}

\textbf{A principled perspective}~We first provide a principled analysis to understand TVM. In a nutshell, we argue that TVM offers effective optimization \emph{directions} that can yield a model under the VRM principle~\cite{NIPS2000_ba9a56ce}.
As aforementioned, task vectors represent cumulative gradients for a specific task. This means that $\tau_t$ and $\tau_s$ express the cumulative gradients produced by training a model on labeled data $\mathcal{D}_t=\{(x_i^t,y_i^t)\}$ and $\mathcal{D}_s=\{(x_j^s,y_j^s)\}$, respectively.
Under certain conditions, the interpolation between $\tau_t$ and $\tau_s$ in \cref{eqn:3} can approximate the gradients produced by training a model on the virtual mix data with a \emph{vicinity distribution}, \ie, $\mathcal{D}_{\text{mix}}=\{(\lambda x_i^t+(1-\lambda)x_j^s,\lambda y_i^t+(1-\lambda)y_j^s)\}$.
Through training on these linearly-interpolated samples, the model learns from a smoother vicinity space and thus leads to a VRM-based model with better generalization abilities, as proven in a multitude of studies~\cite{zhang2018mixup,verma2019manifold,zhang2021does,carratino2022mixup}.
A more formal analysis of TVM is given in Appendix.

\noindent\textbf{An empirical perspective}~To verify the argument above, we further perform empirical studies to analyze the behavior and properties of TVM. Concretely, we construct a paired task vector $(\tau_t,\tau_s)$ from cancer-specific prognostic models $(\mathcal{M}_t,\mathcal{M}_s)$ and show the dynamics of TVM by loss landscape visualization and SAR quantification.
We provide an example here for illustration. Please refer to \cref{sec:5_4_vis} for more results.

\begin{figure}[htbp]
  \centering
   \includegraphics[width=\linewidth]{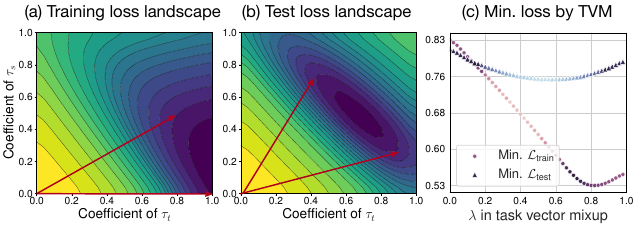}
   \caption{\textbf{Loss landscape of task vector mixup} on $t$. Red vectors in (a) and (b) depict the range of TVM with lower loss. Figure (c) shows the minimum loss that can be reached by TVM.}
   \label{fig:4a_intro_tvm_landscape}
\end{figure}

\begin{figure}[htbp]
  \centering
   \includegraphics[width=0.65\linewidth]{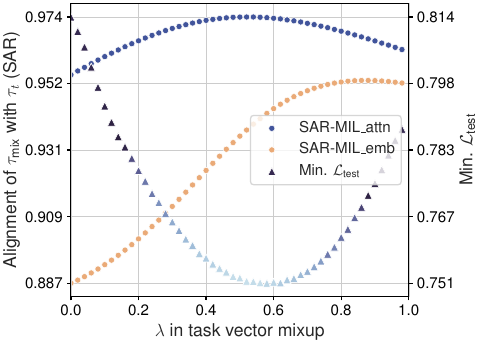}
   \vspace{-.7em}
   \caption{\textbf{Alignment of $\tau_{\text{mix}}$ with $\tau_t$ in dominant subspaces} (95\%) for the two layers of MIL encoder, measured by SAR~\cite{marczakno2025isoc}.}
   \label{fig:4b_intro_tvm_sar}
\end{figure}

\cref{fig:4a_intro_tvm_landscape} shows that TVM can point towards the optimization direction that produces a prognostic model with better generalizability on $t$. For example, there is a prognostic model that obtains a lower loss on both training and test sets when $\lambda\in[0.7,0.8]$.
Furthermore, we analyze the SAR (subspace alignment ratio) of TVM with $\tau_t$ at the two layers of MIL encoder, \ie, $f_{\text{emb}}(\cdot)$ for instance embedding and $f_{\text{attn}}(\cdot)$ for attention-based instance aggregation.
Results in \cref{fig:4b_intro_tvm_sar} show that the TVM of $f_{\text{attn}}(\cdot)$ often keeps a close alignment with $\tau_t$, while that of $f_{\text{emb}}(\cdot)$ does not.
This suggests that the improvements in generalization ability may be attributed more to the attention layer in MIL.

\subsection{Sparse task vector aggregation}
\label{sec:4_3_stva}

The task vector mixture $\tau_{\text{mix}}$ encodes the prognostic knowledge from various cancers. However, it is not always beneficial for $t$, \eg, $\mathcal{M}_{s}$ is not well trained and even performs worse on $s$, or $s$ inherently conflicts with $t$. Besides, $\tau_{\text{mix}}$ could contain redundant knowledge, especially when the prognosis of two cancer diseases is highly correlated. To address these issues, we propose sparse task vector aggregation to integrate the most beneficial knowledge from various, multiple sources, borrowing the idea from Mixture of Experts~\cite{jacobs1991adaptive,shazeer2017outrageously}.

Concretely, for any WSI input $X$, another MIL-based hypernetwork $\mathcal{H}_{\text{agg}}$, which shares its MIL encoder with $\mathcal{H}_{\text{mix}}$ but has an independent output head, is adopted to learn the weights assigned for each task vector mixture, as shown in \cref{fig:3_our_stvm}.
Namely, there is $w=\mathcal{H}_{\text{agg}}(X)=\{w_i\ge 0\mid i\in[1,m]\}$. Similar to $\lambda$, $w$ is also adaptive to inputs, indicating dynamic model weights.
According to $w$, we select the top $K$ mixtures from $\tau_{\text{mix}}$ and then aggregate them into $\tau_{t}^{*}$ by weighted summation, written as
\begin{equation}
    \tau_{t}^{*}=\sum_j w_{j}\tau_{\text{mix},j},
    \label{eqn:4}
\end{equation}
where $j$ is the index of top $K$ elements in $\{w_i\}$ and $\tau_{\text{mix},j}$ is the $j$-th element in $\tau_{\text{mix}}$.

\noindent\textbf{Discussion}~
(1) \textbf{The role of $w$}. As aforementioned, TVM offers possible \emph{directions} towards better model generalization. It may not reach the optimal point solely by adjusting $\lambda$, as shown in \cref{fig:4a_intro_tvm_landscape}. $w$ could help to accomplish this, by optimizing the magnitude of TVM.
(2) \textbf{Comparison with existing methods}. We note that this task vector aggregation scheme is different from existing methods for MTL~\cite{ilharco2023editing,yangadamerging2024,yang2024representation,wang2024localizing,marczakno2025isoc,chen2025fwmergingscalingmodelmerging,wei2025modeling} in terms of both motivation and purpose.
(a) Existing methods generally derive $w$ from a validation set. However, this tends to obtain a biased $w$ in the scenario of WSI prognosis. Because only a few WSI samples can be used, they are prone to deviating from the true data distribution. Given this, we train a hypernetwork to learn input-conditional $w$.
(b) In particular, we impose sparsification on mixed task vectors, in light of the fact that the knowledge offered by some cancer-specific models may be redundant or useless for the cancer of interest.

\subsection{Network training}

As illustrated in \cref{fig:3_our_stvm}, we derive the final target model $\mathcal{M}_t^{*}$ by applying $\tau_{t}^{*}$ to initial model weights, as follows:
\begin{equation}
    \mathcal{M}_{t}^{*}=\mathcal{M}_0+\tau_{t}^{*}.
    \label{eqn:5}
\end{equation}
Unlike traditional network training, here we optimize two hypernetworks, \ie, $\mathcal{H}_{\text{mix}}$ and $\mathcal{H}_{\text{agg}}$, which share an MIL encoder but have independent output heads.
They output suitable $\lambda$ and $w$ for task vector merging.
For any training sample, we adopt a loss function as follows:
\begin{equation}
    \mathcal{L}=\mathcal{L}_{\text{sl}}+\mathcal{L}_{\text{aux}},
    \label{eqn:6}
\end{equation}
where $\mathcal{L}_{\text{sl}}=\mathcal{L}_{\text{NLL}}(y,\hat{y})$ is a negative likelihood loss (NLL)~\cite{zadeh2020bias} frequently-adopted in WSI-based survival analysis for supervised learning.
$\mathcal{L}_{\text{aux}}$ is an auxiliary loss:
\begin{equation}
\begin{aligned}
    \mathcal{L}_{\text{aux}}&=\beta\mathcal{L}_{\text{mix}}+\gamma\mathcal{L}_{\text{agg}}\\
    &=\beta \frac{\sum_j\lambda_j^2}{K}+\gamma(\mathop{\text{log}}{\sum_ie^{w_i}})^2,
    \label{eqn:7}
\end{aligned}
\end{equation}
where $j$ is the index of top $K$ elements in $\{w_i\}$; $\beta$ and $\gamma$ are hyper-parameters controlling loss weights.
$\mathcal{L}_{\text{mix}}$ is an auxiliary loss for mixup coefficients $\lambda$.
It penalizes large values in $\{\lambda_i\}$, utilized to encourage transferring knowledge from $\mathcal{M}_s$.
$\mathcal{L}_{\text{agg}}$ is a penalty loss for aggregating weights $\{w_i\}$.
It can suppress excessively large $w_i$, often adopted in the training of MoE~\cite{zoph2022st}.

\section{Experiments and results}
\label{sec:experiments}

\begin{table*}[htbp]
\centering
\caption{\textbf{Comparison with baselines} in WSI prognosis tasks. It reports the mean and standard deviation of C-Index in five-fold cross-validation. Best in bold, second best underline.}
\label{tab:1_main_cmp_results} 
\tabcolsep=0.1cm
\resizebox{\textwidth}{!}{
\begin{tabular}{l|ccccccccccccc|c}
\toprule
\textbf{Method} & BRCA & KIPAN & LUNG & \makecell[c]{GBM\\ LGG} & \makecell[c]{COAD\\ READ} & STES & UCEC & HNSC & SKCM & BLCA & LIHC & CESC & SARC & \textbf{Avg.} \\
\midrule
\multicolumn{15}{l}{- \textit{Traditional cancer-specific training on} $\mathcal{M}_{t}$} \\ \midrule
& 0.6648 & 0.8094 & 0.5496 & \underline{0.7756} & 0.6725 & 0.6648 & 0.7098 & 0.6201 & 0.5708 & \underline{0.6438} & 0.7265 & 0.6500 & 0.5312 &   \\ 
\multirow{-2}{*}{Vanilla} & ($\pm$ 0.032) & ($\pm$ 0.013) & ($\pm$ 0.034) & ($\pm$ 0.020) & ($\pm$ 0.028) & ($\pm$ 0.040) & ($\pm$ 0.043) & ($\pm$ 0.047) & ($\pm$ 0.037) & ($\pm$ 0.015) & ($\pm$ 0.048) & ($\pm$ 0.058) & ($\pm$ 0.056) & \multirow{-2}{*}{0.6609} \\ 
\multirow{2}{*}{Fine-tuned} & 0.6723 & 0.8080 & 0.5463 & 0.7744 & 0.6753 & 0.6694 & 0.7136 & 0.6207 & 0.5682 & 0.6384 & 0.7253 & 0.6367 & 0.5462 & \multirow{2}{*}{0.6611} \\
& ($\pm$ 0.021) & ($\pm$ 0.014) & ($\pm$ 0.034) & ($\pm$ 0.020) & ($\pm$ 0.028) & ($\pm$ 0.044) & ($\pm$ 0.039) & ($\pm$ 0.043) & ($\pm$ 0.032) & ($\pm$ 0.021) & ($\pm$ 0.046) & ($\pm$ 0.058) & ($\pm$ 0.030) &  \\ \midrule
\multicolumn{15}{l}{- \textit{Representation-based knowledge transfer from $\mathcal{M}_{s}$}} \\ \midrule
\multirow{2}{*}{Fine-tuned} & 0.5121 & 0.6430 & 0.5406 & 0.7713 & 0.6345 & 0.6155 & 0.7146 & 0.5638 & 0.5206 & 0.5971 & 0.6276 & 0.6177 & 0.4764 & \multirow{2}{*}{0.6027} \\ 
& ($\pm$ 0.044) & ($\pm$ 0.050) & ($\pm$ 0.035) & ($\pm$ 0.022) & ($\pm$ 0.039) & ($\pm$ 0.042) & ($\pm$ 0.075) & ($\pm$ 0.052) & ($\pm$ 0.036) & ($\pm$ 0.024) & ($\pm$ 0.056) & ($\pm$ 0.046) & ($\pm$ 0.051) &  \\
& \underline{0.7181} & \underline{0.8096} & \underline{0.5714} & 0.7726 & \textbf{0.7123} & \textbf{0.6708} & \underline{0.7371} & \underline{0.6257} & \textbf{0.5954} & \textbf{0.6644} & \underline{0.7563} & \underline{0.6629} & \underline{0.5596} &  \\
\multirow{-2}{*}{ROUPKT~\cite{liu2025cross}} & ($\pm$ 0.051) & ($\pm$ 0.007) & ($\pm$ 0.033) & ($\pm$ 0.019) & ($\pm$ 0.047) & ($\pm$ 0.042) & ($\pm$ 0.054) & ($\pm$ 0.034) & ($\pm$ 0.025) & ($\pm$ 0.017) & ($\pm$ 0.048) & ($\pm$ 0.061) & ($\pm$ 0.040) & \multirow{-2}{*}{\underline{0.6812}}  \\ \midrule
\multicolumn{15}{l}{- \textit{Model merging-based knowledge transfer from $\mathcal{M}_{s}$}} \\ \midrule
\multirow{2}{*}{Model Avg.} & 0.5450 & 0.5849 & 0.5306 & 0.7401 & 0.6260 & 0.5707 & 0.6971 & 0.5670 & 0.4812 & 0.5843 & 0.5553 & 0.6179 & 0.4448 & \multirow{2}{*}{0.5804} \\ 
& ($\pm$ 0.066) & ($\pm$ 0.047) & ($\pm$ 0.037) & ($\pm$ 0.021) & ($\pm$ 0.050) & ($\pm$ 0.052) & ($\pm$ 0.082) & ($\pm$ 0.067) & ($\pm$ 0.026) & ($\pm$ 0.028) & ($\pm$ 0.058) & ($\pm$ 0.051) & ($\pm$ 0.065) &  \\
\multirow{2}{*}{AdaMerging~\cite{yangadamerging2024}} & 0.5534 & 0.5529 & 0.5260 & 0.7477 & 0.5971 & 0.5530 & 0.6791 & 0.5384 & 0.4812 & 0.5928 & 0.5567 & 0.5826 & 0.4341 & \multirow{2}{*}{0.5689} \\ 
& ($\pm$ 0.051) & ($\pm$ 0.044) & ($\pm$ 0.036) & ($\pm$ 0.020) & ($\pm$ 0.047) & ($\pm$ 0.043) & ($\pm$ 0.083) & ($\pm$ 0.067) & ($\pm$ 0.040) & ($\pm$ 0.025) & ($\pm$ 0.052) & ($\pm$ 0.061) & ($\pm$ 0.034) &  \\
\multirow{2}{*}{TIES AM~\cite{yadav2023ties}} & 0.6799 & 0.7866 & 0.5319 & 0.7699 & 0.6444 & 0.6080 & 0.7174 & 0.5522 & 0.5781 & 0.6401 & 0.7009 & 0.6418 & 0.4637 & \multirow{2}{*}{0.6396} \\ 
& ($\pm$ 0.037) & ($\pm$ 0.015) & ($\pm$ 0.027) & ($\pm$ 0.018) & ($\pm$ 0.029) & ($\pm$ 0.052) & ($\pm$ 0.026) & ($\pm$ 0.046) & ($\pm$ 0.026) & ($\pm$ 0.043) & ($\pm$ 0.062) & ($\pm$ 0.059) & ($\pm$ 0.035) &  \\
\multirow{2}{*}{Surgery AM~\cite{yang2024representation}} & 0.5448 & 0.7172 & 0.5335 & 0.7696 & 0.6137 & 0.5825 & 0.6964 & 0.5670 & 0.4933 & 0.5881 & 0.5794 & 0.6067 & 0.4342 & \multirow{2}{*}{0.5943} \\ 
& ($\pm$ 0.052) & ($\pm$ 0.035) & ($\pm$ 0.038) & ($\pm$ 0.023) & ($\pm$ 0.049) & ($\pm$ 0.059) & ($\pm$ 0.076) & ($\pm$ 0.066) & ($\pm$ 0.040) & ($\pm$ 0.022) & ($\pm$ 0.055) & ($\pm$ 0.062) & ($\pm$ 0.065) &  \\
\multirow{2}{*}{Iso-C AM~\cite{marczakno2025isoc}} & 0.5425 & 0.7058 & 0.5499 & 0.7253 & 0.5532 & 0.4753 & 0.7334 & 0.4946 & 0.4634 & 0.5862 & 0.5406 & 0.5131 & 0.5252 & \multirow{2}{*}{0.5699} \\ 
& ($\pm$ 0.056) & ($\pm$ 0.033) & ($\pm$ 0.042) & ($\pm$ 0.016) & ($\pm$ 0.044) & ($\pm$ 0.080) & ($\pm$ 0.067) & ($\pm$ 0.064) & ($\pm$ 0.043) & ($\pm$ 0.075) & ($\pm$ 0.084) & ($\pm$ 0.109) & ($\pm$ 0.026) &  \\
& \textbf{0.7408} & \textbf{0.8228} & \textbf{0.5836} & \textbf{0.7763} & \underline{0.7069} & \underline{0.6698} & \textbf{0.7830} & \textbf{0.6569} & \underline{0.5948} & 0.6413 & \textbf{0.7585} & \textbf{0.6965} & \textbf{0.6024} &  \\
\multirow{-2}{*}{\textbf{$\ours$}} & ($\pm$ 0.060) & ($\pm$ 0.015) & ($\pm$ 0.042) & ($\pm$ 0.022) & ($\pm$ 0.029) & ($\pm$ 0.050) & ($\pm$ 0.072) & ($\pm$ 0.042) & ($\pm$ 0.027) & ($\pm$ 0.038) & ($\pm$ 0.036) & ($\pm$ 0.065) & ($\pm$ 0.070) & \multirow{-2}{*}{\textbf{0.6949}} \\
\bottomrule
\end{tabular}
}
\end{table*}

\subsection{Experimental settings}

\textbf{Datasets}~Following \cite{liu2025cross}, 13 cancer datasets from TCGA, each of which corresponds to one cancer type, are used in experiments. They cover 8,818 diagnostic WSIs from 7,268 patients. WSIs are processed into patch features by UNI~\cite{chen2024towards}. The number of patients in each dataset ranges from 248 (TCGA-SARC) to 1,035 (TCGA-BRCA), with an overall disease-specific survival (DSS) rate of 76.82\%. Each dataset includes 5-fold cross-validation splits.

\noindent\textbf{Baselines}~Three categories of baselines are compared in experiments.
(1) Traditional cancer-specific training. It contains vanilla and fine-tuned prognostic models, where ``fine-tuned" refers to fine-tuning the last prediction layer of $\mathcal{M}_t$ while keeping the others frozen.
(2) Representation-based knowledge transfer from $\mathcal{M}_s$: ``fine-tuned" means fine-tuning the last prediction layer of $\mathcal{M}_s$ on $t$ while freezing the WSI-level representations output by $\mathcal{M}_s$ where $\mathcal{M}_s$ is a model with the best transfer performance on $t$; ROUPKT~\cite{liu2025cross} is a routing-based network that utilizes the representations output by $\{\mathcal{M}_{s_i}\}$.
(3) Model merging-based knowledge transfer from $\mathcal{M}_s$. Besides model average, existing generic model merging methods are compared: AdaMerging (AM)~\cite{yangadamerging2024}, TIES~\cite{yadav2023ties}, Surgery~\cite{yang2024representation}, and Iso-C~\cite{marczakno2025isoc}. Other recent methods are not presented as they are tailored for MTL or a specific architecture.
AM is a general method for optimizing the $w$ in $\tau_{\text{new}}=w_t\tau_t+\sum_iw_{s_i}\tau_{s_i}$. To adapt it to this task, the same $\mathcal{L}_{\text{sl}}$ is employed to train $w$.
The other three baselines are adopted to merge each paired $\tau_t$ and $\tau_{s_i}$.
AM is also used along with them to learn the aggregating weight $w$.

\noindent\textbf{Implementation details}~All cancer-specific prognostic models, \ie, $\mathcal{M}_t$ and $\{\mathcal{M}_{s_i}\}_{i=1}^m$ where $m=12$, are implemented by a general attention-based MIL architecture~\cite{ilse2018attn}, following the training recipe from \cite{pmlr-v235-song24b}.
For other compared models, the same training settings are used.
In $\ours$, the hypernetworks, $\mathcal{F}_{\text{mix}}$ and $\mathcal{F}_{\text{agg}}$, share a simple mean-MIL encoder, each of which has an independent fully-connected layer as its output head.
For hyper-parameters, $K$ is set to 5 and the coefficient of $\mathcal{L}_{\text{mix}}$, $\beta$, is set to 0.05 by default across all datasets; the coefficient of $\mathcal{L}_{\text{agg}}$, $\gamma$, is derived by cross-validation on training samples.
For performance evaluation, a frequently adopted metric, C-Index, is measured; we report its mean and standard deviation in 5-fold cross-validation.
For more details on experimental settings, please refer to Appendix.

\subsection{Comparison with baselines}

Comparative results are shown in \cref{tab:1_main_cmp_results}. We have three main observations as follows.
(1) $\ours$ outperforms cancer-specific learning by 5.14\% on average and obtains better performance on 12 out of 13 datasets. This suggests that $\ours$ almost always enhances the target model further through effectively utilizing the knowledge from other cancers.
(2) $\ours$ surpasses the representation-based knowledge transfer baseline on 9 out of 13 datasets, with an overall improvement of 2.01\%. Moreover, $\ours$ has significantly lower inference costs, as exhibited in \cref{fig:2_perf_overview}. These head-to-head comparisons confirm the superiority of our scheme over the representation-based solution in terms of both computation efficiency and predictive performance.
(3) Existing model merging approaches often struggle to leverage the knowledge from other cancers for a target, while our $\ours$ does not. Because existing approaches generally consider developing multi-task capabilities by resolving task interference, while $\ours$ aims to boost the model performance on a single task by utilizing the most beneficial, generalizable knowledge from others.

\subsection{Visualization results}
\label{sec:5_4_vis}

\begin{figure}[htbp]
  \centering
   \includegraphics[width=\linewidth]{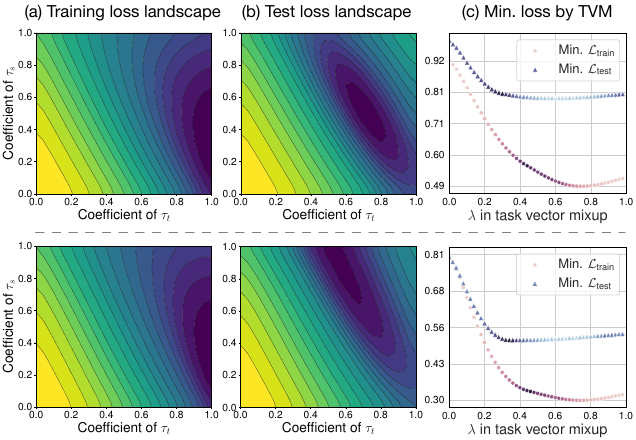}
   \caption{\textbf{More results of TVM loss landscape}.}
   \label{fig:5a_more_tvm_landscape}
\end{figure}

\begin{figure}[htbp]
  \centering
   \includegraphics[width=\linewidth]{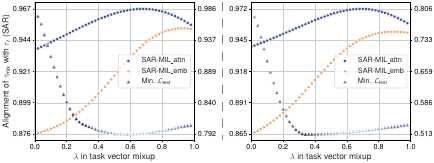}
   \caption{\textbf{More results of the alignment of $\tau_{\text{mix}}$ with $\tau_t$}.}
   \label{fig:5b_more_tvm_sar}
\end{figure}

\textbf{More empirical results of TVM}~As shown in \cref{fig:5a_more_tvm_landscape} and \cref{fig:5b_more_tvm_sar}, we observe similar results on other task vector pairs: \textit{i}) TVM offers possible optimization directions towards better model generalization and \textit{ii}) $\tau_{\text{mix}}$ often aligns with $\tau_t$ when applied to the attention layer of MIL encoder.

\begin{figure}[bp]
  \centering
   \includegraphics[width=\linewidth]{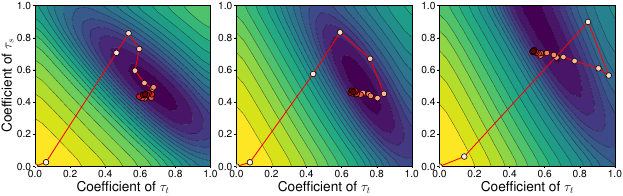}
   \caption{\textbf{Visualization of TVM trajectory} in training. $\ours$ is used to train the paired $\tau_t$ and $\tau_s$ (same as those in \cref{fig:4a_intro_tvm_landscape} and \cref{fig:5a_more_tvm_landscape}). A deeper point color indicates a larger epoch number.}
   \label{fig:6_tvm_traj}
\end{figure}

\noindent\textbf{Trajectory of TVM in optimization}~We further visualize the trajectory of TVM over the landscape of test loss in training, to examine how TVM changes and whether it can step into the area with smaller test loss on $t$.
For a better view on 2D plane, a pair of $\tau_t$ and $\tau_{s}$ is also constructed.
We train its mixture using $\ours$ and record the value of $\lambda$ and $w$ (averaged over training samples) at each epoch to visualize the location of task vector mixture in the landscape of test loss. Visualization results are provided in \cref{fig:6_tvm_traj}.
These results further imply that our $\ours$ can effectively improve the generalization ability of cancer-specific models by utilizing the knowledge of other cancers.

\begin{figure}[htbp]
  \centering
   \includegraphics[width=\linewidth]{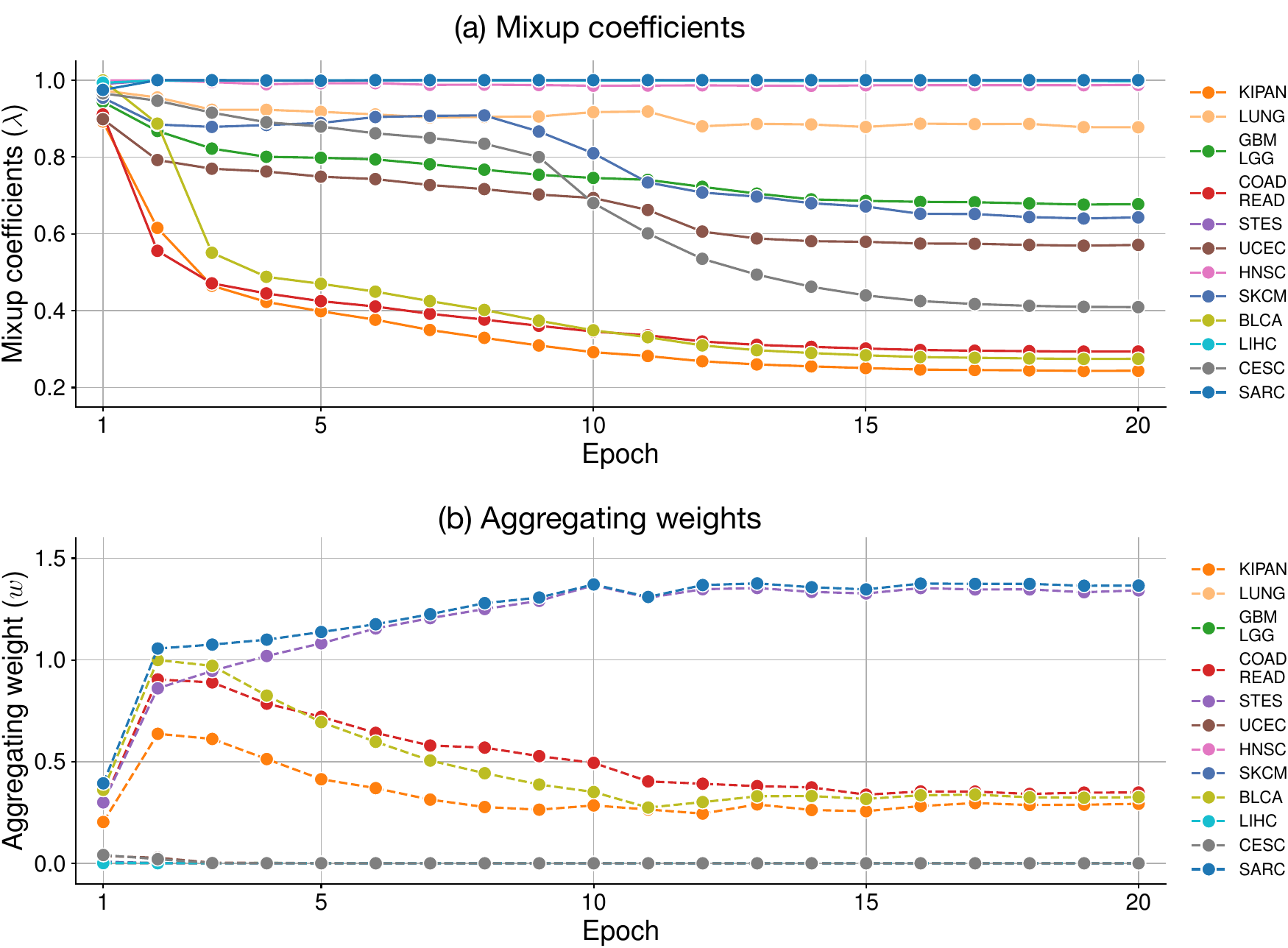}
   \caption{\textbf{Dynamics of $\lambda$ and $w$} in training. This example shows a case where $t=$ BRCA and 12 cancer-specific models ($\mathcal{M}_s$) are merged with $\mathcal{M}_t$ by $\ours$ to obtain $\mathcal{M}_t^{*}$.}
   \label{fig:7_stvm_weight}
\end{figure}

\noindent\textbf{Dynamics of $\lambda$ and $w$ in training}~
Here we investigate a real case ($t=$ BRCA and $m=12$), instead of a model pair (\ie, $m=1$), to study the optimization dynamics of $\lambda$ and $w$.
As shown in \cref{fig:7_stvm_weight}, we find that \textit{i}) two models have $\lambda_i=1$ while associated with larger $w_i$ and \textit{ii}) there are three models of other cancers (KIPAN, COADREAD, and BLCA) that exhibit lower $\lambda_i$ ($<0.3$) in TVM and relatively large $w_i$ in aggregation.
This means that $\mathcal{M}_{\text{BRCA}}$ is encouraged to leverage prognostic knowledge from KIPAN, COADREAD, and BLCA. This observation indicates that cross-cancer prognostic models are utilized in the learning process. These models transferred from other cancers help $\mathcal{M}_{\text{BRCA}}$ perform better in WSI prognosis, reflected by the 11.4\% improvement on BRCA in \cref{tab:1_main_cmp_results}.

\subsection{Ablation study}

We perform ablation studies on two key components of $\ours$, \ie, task vector mixup and sparse mixture aggregation. Their results are exhibited in \cref{tab:2_abl}.
(1) For task vector mixup, we observe that \textit{i}) there are performance drops when no interpolation between $\tau_t$ and $\tau_{s_i}$ is used and \textit{ii}) hypernetwork-driven $\lambda_i$ yields a slight improvement in overall performance over its trainable-parameter-based variant.
(2) For sparse mixture aggregation, we find that hypernetwork-driven $w$ is crucial, compared to a unified $w$ for all inputs. This result indicates the importance of dynamic $w$ for different WSI samples. As mentioned in \cref{sec:4_3_stva}, although TVM offers better optimization directions, it is $w$ that ultimately determines the position of TVM and plays a critical role in finding optimal models.
Moreover, the model benefits from the sparsity in $\tau_{\text{mix}}$.
These results verify the effectiveness of our core designs.

\begin{table}[htbp]
\centering
\caption{\textbf{Ablation study} on $\ours$.}
\label{tab:2_abl} 
\resizebox{0.80\linewidth}{!}{
\begin{tabular}{llc}
\toprule
\multicolumn{2}{c}{\textbf{Ablation}} & \textbf{Avg.} \\
\midrule
\multicolumn{2}{l}{- \textit{On task vector mixup}} \\ \midrule
\multirow{3}{*}{w/o mixup} & fix $\lambda$ = 0 (only $\{\tau_{s_i}\}$)  & 0.6860 \\
& fix $\lambda$ = 0 ($\tau_t\in\{\tau_{s_i}\}$)  & 0.6895 \\
 & fix $\lambda$ = 1 (only $\tau_t$) & 0.6851 \\ \cmidrule(r){1-2}
\multirow{2}{*}{w/  mixup} & trainable param. $\lambda$ & 0.6921 \\
 & hypernetwork-driven $\lambda$ & \textbf{0.6949} \\ \midrule
\multicolumn{2}{l}{- \textit{On task vector aggregating}} \\ \midrule
w/o sparsity & hypernetwork-driven $w$ & 0.6912 \\ \cmidrule(r){1-2}
\multirow{2}{*}{w/  sparsity} & trainable param. $w$ & 0.6490 \\
 & hypernetwork-driven $w$ & \textbf{0.6949} \\
\bottomrule
\end{tabular}
}
\end{table}

\subsection{More experiments and analysis}

\textbf{Hypernetwork-driven aggregation}~An adaptive $w$ is important for task vector aggregation, as discussed in \cref{sec:4_3_stva} and shown in \cref{tab:2_abl}. To further examine its potential, we incorporate it into existing model merging methods.
From the results given in \cref{tab:3_bsl_w_ours}, we find that our scheme can boost the performance of existing methods by a large margin (14.5\% on average).
This confirms the importance of input-conditional weights for this task.
Intuitively, compared with a unified $w$, the input-conditional $w$ enables a flexible model that can adjust its strength of connecting with $\mathcal{M}_{s_i}$ according to the current input.

\begin{table}[htbp]
\centering
\caption{\textbf{Our hypernetwork-driven aggregation scheme} for existing model merging methods.}
\label{tab:3_bsl_w_ours}
\resizebox{0.9\linewidth}{!}{
\begin{tabular}{llc}
\toprule
\multicolumn{2}{c}{\textbf{Method}} & \textbf{Avg.} \\
\midrule
AdaMerging &  vanilla (w/ trainable param. $w$) & 0.5689 \\
AdaMerging &  w/ our hypernetwork-driven $w$  & \textbf{0.6877} \\ \cmidrule(r){1-2}
TIES &  w/ AM & 0.6396 \\
TIES &  w/ our hypernetwork-driven $w$  & \textbf{0.6802} \\ \cmidrule(r){1-2}
Surgery & w/ AM  & 0.5943 \\
Surgery &  w/ our hypernetwork-driven $w$  & \textbf{0.6668} \\ \cmidrule(r){1-2}
Iso-C &  w/ AM & 0.5699 \\
Iso-C  &  w/ our hypernetwork-driven $w$  & \textbf{0.6761} \\
\bottomrule
\end{tabular}
}
\end{table}

\noindent\textbf{Hyper-parameters}~We evaluate $\ours$ with different hyper-parameter settings. \cref{tab:4_hyper_param_beta} and \cref{tab:5_hyper_param_gamma} show that it is suitable to impose mild penalties on $\lambda$ and $w$.

\begin{table}[htbp]
\centering
\caption{\textbf{$\ours$ with different $\beta$} (coefficient of $\mathcal{L}_{\text{mix}}$).}
\label{tab:4_hyper_param_beta}
\resizebox{0.65\linewidth}{!}{
\begin{tabular}{lcccc}
\toprule
$\beta$ & $0$ & $0.01$ & $0.05$ & $0.1$ \\
\midrule
\textbf{Avg.} & 0.6927 & 0.6933 & \textbf{0.6949} & 0.6909 \\
\bottomrule
\end{tabular}
}
\end{table}

\begin{table}[htbp]
\centering
\caption{\textbf{$\ours$ with different $\gamma$} (coefficient of $\mathcal{L}_{\text{agg}}$).}
\label{tab:5_hyper_param_gamma}
\resizebox{0.8\linewidth}{!}{
\begin{tabular}{lccccc}
\toprule
$\gamma$ & $0$ & $0.0025$ & $0.005$ & $0.01$ & 0.05 \\
\midrule
\textbf{Avg.} & 0.6879 & \textbf{0.6896} & 0.6896 & 0.6865 & 0.6881 \\
\bottomrule
\end{tabular}
}
\end{table}

\noindent\textbf{Additional experiments}~For more experiments, \eg, visualization, hyper-parameter $K$, comparison with MIL networks, and training efficiency, please refer to Appendix.

\section{Discussion}
\label{sec:discussion}

Classical cancer-specific learning paradigm faces critical challenges posed by limited data in histopathology and high heterogeneity in tumors.
Although better WSI representation approaches have been invented, they still struggle to learn generalizable prognostic knowledge from small data.
Cross-cancer knowledge transfer, as a new learning paradigm, has shown potential in addressing this issue.
Following this line of research, this paper presents an effective scheme for efficient prognosis knowledge transfer.
Nevertheless, there are some limitations in this study.
We discuss them and future works as follows.
(\textbf{1}) The datasets this study relies on are from TCGA. Although covering diverse cancer types, the number of patients is still very limited for certain types, \eg, cervical and liver cancer ($N<400$). This poses great challenges to model evaluation. It would be better to adopt repeated cross-validation with more folds when additional computation overhead is allowed.
(\textbf{2}) Our experiments are based on a general MIL network architecture. More advanced architectures could be further adopted to evaluate algorithms.
(\textbf{3}) $\ours$ still requires access to training data to obtain suitable merging weights. It would be exciting to study training-free strategies for this task, like most existing studies on model merging.  

\section{Conclusion}
\label{sec:conclusion}

This paper presents Sparse Task Vector Mixup with Hypernetworks ($\ours$) to efficiently absorb prognostic knowledge from other cancers to improve the generalization performance of cancer-specific models.
Our principled analysis and empirical studies imply that task vector mixup helps to improve the performance of target models by offering directions towards better model generalization.
Extensive experiments show that $\ours$ performs better than traditional cancer-specific learning on 12 out of 13 datasets, with an overall improvement of 5.14\%.
Compared to an existing knowledge transfer solution, our scheme often exhibits better predictive performances yet significantly lower computational costs.
These impressive results suggest that $\ours$ could provide an effective approach for efficient knowledge transfer in WSI prognosis.

\section*{Acknowledgements}

This work was supported by the National Natural Science Foundation of China (grant nos. U22A2037, 62425204, 62122025, 62450002, 62432011).
{
    \small
    \bibliographystyle{ieeenat_fullname}
    \bibliography{main}
}

\clearpage

\begin{appendix}

\setcounter{page}{1}
\setcounter{section}{0}
\setcounter{figure}{0}
\setcounter{table}{0}
\setcounter{equation}{0}

\renewcommand{\figurename}{Figure}
\renewcommand{\thefigure}{S\arabic{figure}}
\renewcommand{\tablename}{Table}
\renewcommand{\thetable}{S\arabic{table}}

\maketitlesupplementary

\section{Derivations of subspace alignment ratio}

Here, we complete the derivation of subspace alignment ratio (SAR), following \cite{marczakno2025isoc}. Recall the definition of SAR given in the main paper, as follows:
\begin{equation}
    \text{SAR}(\tau_{t},\tau_{\text{new}})=\frac{\|\text{Proj}_{\tau_{\text{new}}^{\alpha}}\tau_{t}\|_F}{\|\tau_{t}\|_F}.
    \label{eqn:s_1}
\end{equation}
$\tau_{\text{new}}^{\alpha}$ is the subspace spanned by the top $\alpha$ left-singular vectors of $\tau_{\text{new}}$. It represents a dominant subspace of $\tau_{\text{new}}$, expressed as follows:
\begin{equation}
  \tau_{\text{new}}^{\alpha}=U_{\text{new}}^{\alpha}(U_{\text{new}}^{\alpha})^{{\top}}.
    \label{eqn:s_2}
\end{equation}
$U_{\text{new}}^{\alpha}$ corresponds to the first $R_{\alpha}$ columns of $U_{\text{new}}$.
$U_{\text{new}}$ is obtained from the SVD (Singular Value Decomposition) decomposition of $\tau_{\text{new}}$, \ie, 
\begin{equation}
\tau_{\text{new}}=U_{\text{new}}\sum (V_{\text{new}})^{\top},
\label{eqn:s_3}
\end{equation}
where $\sum=\text{diag}(\sigma_1,\cdots,\sigma_R)$ contains the singular values of $\tau_{\text{new}}$.
$R_{\alpha}$ is calculated by
\begin{equation}
\begin{aligned}
R_{\alpha} &= \min \left\{ R_{\alpha} : \| \tau_{\text{new}} - \tau_{\text{new}}^{\alpha} \|_F \leq (1-\alpha) \| \tau_{\text{new}}^{\alpha} \|_F \right\}\\
&= \min \left\{ R_{\alpha} : \frac{\sum_{r=R_{\alpha}+1}^R \sigma_r^2}{\sum_{r=1}^R \sigma_r^2} \leq (1-\alpha)^2 \right\}.
\label{eqn:s_4}
\end{aligned}
\end{equation}
This ensures that the approximation error between $\tau_{\text{new}}^{\alpha}$ and $\tau_{\text{new}}$ is equal to or less than $1-\alpha$. Therefore, $\tau_{\text{new}}^{\alpha}$ indicates the dominant subspace of $\tau_{\text{new}}$ (\eg, $\alpha=0.95$).
$\text{Proj}_{\tau_{\text{new}}^{\alpha}}\tau_{t}$ is the projection of $\tau_{t}$ onto $\tau_{\text{new}}^{\alpha}$, calculated by
\begin{equation}
\text{Proj}_{\tau_{\text{new}}^{\alpha}}\tau_{t}=U_{\text{new}}^{\alpha}(U_{\text{new}}^{\alpha})^{{\top}}\tau_t.
\label{eqn:s_5}
\end{equation}

\section{A formal analysis of task vector mixup}

Let $\mathcal{M}_t$ and $\mathcal{M}_s$ denote two independent models, initialized by random model weights $\mathcal{M}_0$ and trained on two disjoint datasets $\mathcal{D}_t=\{(x_i^t,y_i^t)\}_{i=1}^{N_t}$ and $\mathcal{D}_s=\{(x_j^s,y_j^s)\}_{j=1}^{N_s}$ by optimizing a supervised loss function $\mathcal{L}$ with the SGD (Stochastic Gradient Descent) algorithm, respectively.

From an optimization perspective, task vectors are the sum of the cumulative gradient $\nabla$ contributed by each labeled sample in iterative training, \ie,
\begin{equation}
\begin{aligned}
\tau_{t}&=\mathcal{M}_t-\mathcal{M}_0=\sum_{i=1}^{N_t}\nabla\mathcal{L}(\mathcal{M}(x_i^t),y_i^t),\\
\tau_{s}&=\mathcal{M}_s-\mathcal{M}_0=\sum_{j=1}^{N_s}\nabla\mathcal{L}(\mathcal{M}(x_j^s),y_j^s).
\label{eqn:s_6}
\end{aligned}
\end{equation}
We write an interpolation between $\tau_{t}$ and $\tau_{s}$ as follows:
\begin{equation}
\tau_{\text{mix}} = \lambda\tau_{t} + (1-\lambda)\tau_{s},
\label{eqn:s_7}
\end{equation}
where $\lambda\in[0,1]$.
Thus, we have
\begin{equation}
\begin{aligned}
\tau_{\text{mix}} &= \lambda\sum_i\nabla\mathcal{L}(\mathcal{M}(x_i^t),y_i^t) + (1-\lambda)\sum_j\nabla\mathcal{L}(\mathcal{M}(x_j^s),y_j^s),\\
&=\frac{\sum_i\sum_j\left(\lambda\nabla\mathcal{L}(\mathcal{M}(x_i^t),y_i^t)+(1-\lambda)\nabla\mathcal{L}(\mathcal{M}(x_j^s),y_j^s)\right)}{N_t*N_s}.
\label{eqn:s_8}
\end{aligned}
\end{equation}
Let $\hat{y}=\mathcal{M}(x)$. When $\frac{\partial\mathcal{L}}{\partial\hat{y}}$ is a convex function \textit{w.r.t.} $\hat{y}$, there is
\begin{equation}
\begin{aligned}
&\lambda\nabla\mathcal{L}(\hat{y}_i^t,y_i^t)+(1-\lambda)\nabla\mathcal{L}(\hat{y}_j^s),y_j^s)\\
&\ge\nabla\mathcal{L}\left(\lambda\hat{y}_i^t+(1-\lambda)\hat{y}_j^s,\lambda y_i^t+(1-\lambda)y_j^s\right).
\label{eqn:s_9}
\end{aligned}
\end{equation}
Let $h=\mathcal{M}_{E}(x)$ where $\mathcal{M}_{E}$ is the MIL encoder of $\mathcal{M}$.
Since the last layer of $\mathcal{M}$ is usually a fully-connected layer for prediction, we further have
\begin{equation}
\begin{aligned}
&\lambda\nabla\mathcal{L}(\hat{y}_i^t,y_i^t)+(1-\lambda)\nabla\mathcal{L}(\hat{y}_j^s),y_j^s)\\
&\ge\nabla\mathcal{L}\left(\lambda h_i^t+(1-\lambda)h_j^s,\lambda y_i^t+(1-\lambda)y_j^s\right)
\label{eqn:s_10}
\end{aligned}
\end{equation}
This means that under certain conditions $\tau_{\text{mix}}$ can be cast (approximately) as the cumulative gradients obtained by training on the interpolation of any paired \textit{MIL representation} $(h_i^t, h_j^s)$. Although this interpolation does not directly lead to that applied to inputs $(x_i^t, x_j^s)$ due to the nonlinearity of $\mathcal{M}$, existing studies~\cite{verma2019manifold,carratino2022mixup} have shown that representation interpolation helps to train a VRM-based model with better generalization ability.

\section{More details on experimental settings}

\subsection{Datasets}
\label{sec:c_1}

There are 13 WSI datasets used in experiments. We show their characteristics in \cref{tab:s_1_data}. Each dataset is split into 5 folds for cross-validation in algorithm evaluation, derived from \cite{liu2025cross}.
Histopathology WSIs are processed by UNI2-h~\cite{chen2024towards}.
Specifically, for one image, tissue regions are segmented and tiled into image patches with 256 $\times$ 256 pixels at 20$\times$ magnification, followed by patch feature extraction using UNI2-h.
The feature extraction converts each patch into a 1,536-dimensional feature vector.

\begin{table}[htbp]
\centering
\caption{\textbf{Characteristics of WSI datasets}.}
\label{tab:s_1_data}
\tabcolsep=0.12cm
\resizebox{\linewidth}{!}{
\begin{tabular}{lrrrrc}
\toprule
\textbf{Dataset} & \textbf{\# Patients} & \textbf{\# WSIs} & \makecell[c]{\textbf{\# Patches}\\ (\textbf{avg.})} & \makecell[c]{\textbf{\# Patches}\\ (\textbf{max})} & \makecell[c]{\textbf{Survival}\\ \textbf{rate}} \\
\midrule
BRCA & 1,035 & 1,106 & 10,694 & 58,176 & 92.4\% \\
KIPAN & 880 & 910 & 10,117 & 72,795 & 83.6\% \\
LUNG & 848 & 924 & 13,199 & 220,294 & 78.2\% \\
GBMLGG & 841 & 1,623 & 16,209 & 216,885 & 52.4\% \\
COADREAD & 568 & 577 & 6,458 & 38,322 & 86.6\% \\
STES & 512 & 539 & 8,429 & 31,458 & 72.7\% \\
UCEC & 504 & 565 & 14,203 & 73,141 & 89.3\% \\
HNSC & 427 & 449 & 6,530 & 56,558 & 71.7\%  \\
SKCM & 412 & 453 & 9,593 & 136,934 & 60.7\% \\
BLCA & 372 & 443 & 14,465 & 103,807 & 68.5\% \\
LIHC & 355 & 362 & 8,875 & 35,839 & 77.5\% \\
CESC & 266 & 276 & 6,212 & 54,021 & 81.6\% \\
SARC & 248 & 591 & 27,346 & 243,906 & 68.5\% \\
\bottomrule
\end{tabular}
}
\end{table}

\subsection{Implementation details}
\label{sec:c_2}

\subsubsection{Model implementation and training}

For any prognostic model obtained by traditional cancer-specific training, \ie, $\mathcal{M}_t$ or any $\mathcal{M}_s$, its implementation and network training follow \citet{pmlr-v235-song24b}.

Specifically, the network is implemented by an ABMIL architecture with an MLP as its instance embedding layer $f_{\text{emb}}(\cdot)$, a gated attention layer $f_{\text{attn}}(\cdot)$ for multi-instance aggregation, and a fully-connected layer as its prediction head at the end.
Each model is trained under the following settings: 20 epoch numbers, a learning rate of 0.0001 with a cosine annealing schedule (the epoch number of warming up is set to 1), an optimizer of AdamW with a weight decay of 0.00001, and a batch size of 1 (one bag) with 16 bags for gradient accumulation, and a frequently-adopted NLL loss for supervised learning in survival modeling~\citep{zadeh2020bias}.
For the other models presented in \cref{tab:1_main_cmp_results}, their training settings are identical to those stated above. All experiments are run on two NVIDIA GeForce RTX 3090 GPUs.

For knowledge transfer methods, \ie, representation-based and model merging-based baselines, the $\mathcal{M}_s$ used in experiments is a prognostic model trained on the first data split in $\mathcal{D}_s$; $\mathcal{M}_t$ is from the data split corresponding to current training fold to prevent data leakage.
For our $\ours$, the hyper-parameter $\gamma$ is derived by a grid search over $\{0, 0.0025, 0.005, 0.01, 0.05\}$ through cross-validation on training samples. It is set to 0.005 for the dataset with over 500 patients and 0.05 for the others.
The other hyper-parameters, $K$ and $\beta$, are set to default values across all datasets without hyper-parameter searching, as stated in the main paper.
For more details, please refer to our source code (see Supplementary File).

\subsubsection{Efficiency evaluation}

In \cref{fig:2_perf_overview}, we evaluate the inference overhead of different models using a bag of 10,000 instances. Note that a bag of 10,000 instances is a general case for gigapixel WSIs, as shown in \cref{tab:s_1_data}.

\subsubsection{Loss landscape visualization}

For $\tau_s$ and $\tau_t$, we uniformly sample task vector coefficients (denoted as $C_s$ and $C_t$) from $[0,1]$ with a step size of 0.04.
Given a group of coefficients $(C_s, C_t)$, we obtain a merged model by $\mathcal{M}_{\text{new}}=\mathcal{M}_0 +C_s\tau_s+C_t\tau_t$ and then measure its $\mathcal{L}_{\text{test}}$ on $t$.
Based on the test loss under different coefficient combinations, we generate a full loss landscape using a Gaussian smooth with $\sigma=1.0$.

\section{Additional experimental results}

\subsection{Dynamics of $\lambda$ and $w$ in training}

\begin{figure}[htbp]
  \centering
   \includegraphics[width=\linewidth]{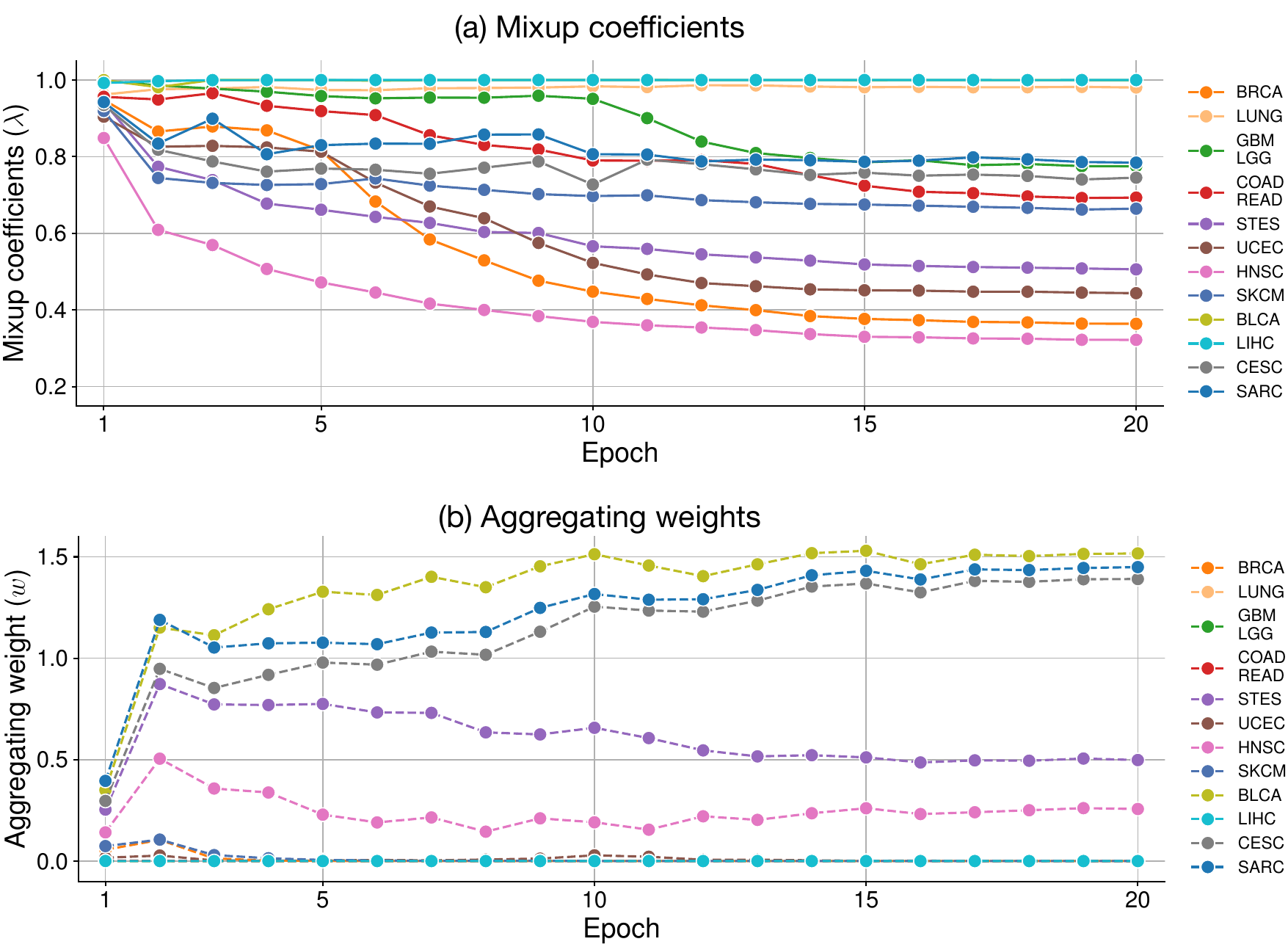}
   \caption{\textbf{Dynamics of $\lambda$ and $w$ in training} ($t=$ KIPAN).}
   \label{fig:s_1_kipan_stvm_weight}
\end{figure}

\begin{figure}[htbp]
  \centering
   \includegraphics[width=\linewidth]{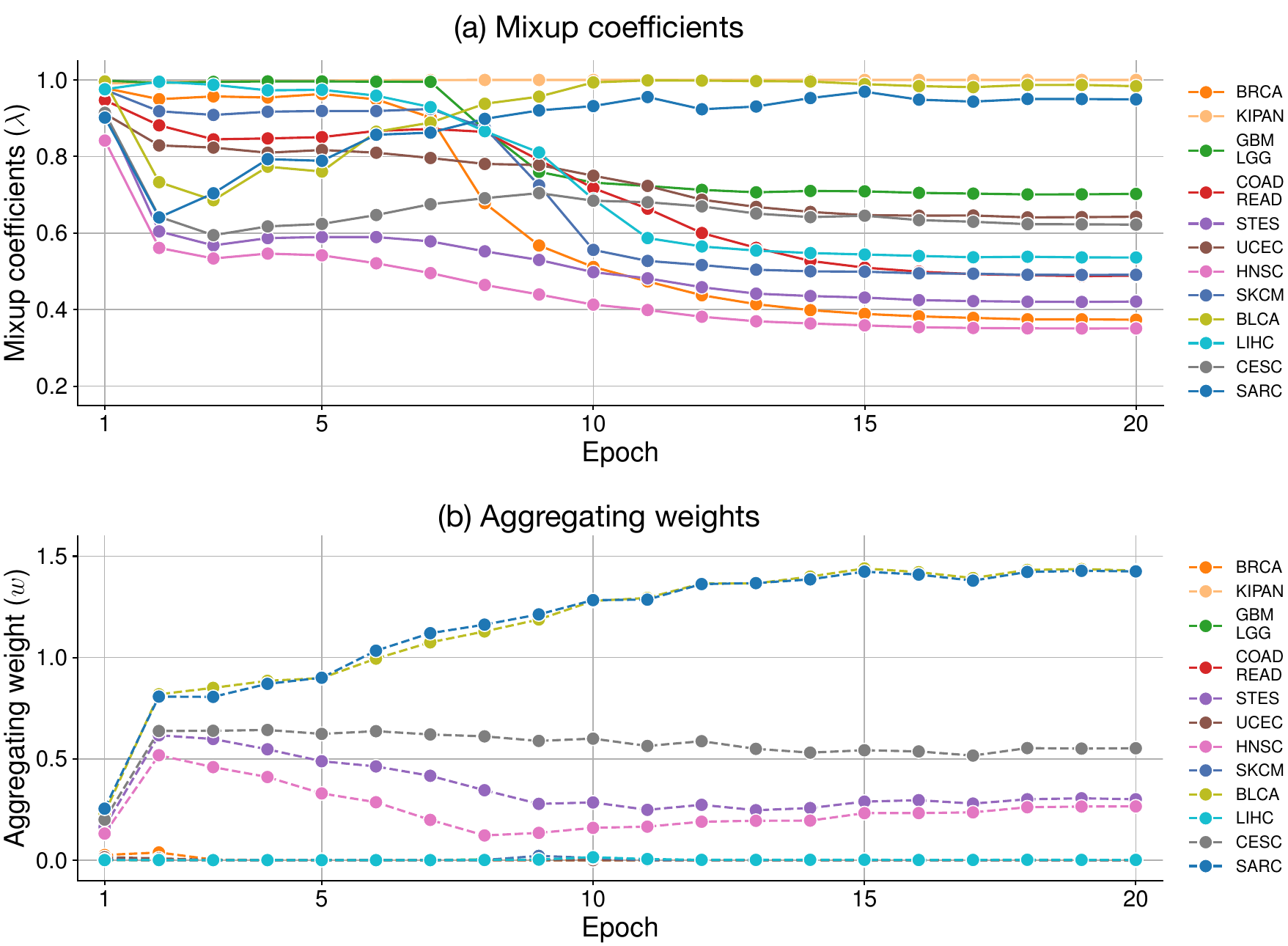}
   \caption{\textbf{Dynamics of $\lambda$ and $w$ in training} ($t=$ LUNG).}
   \label{fig:s_2_lung_stvm_weight}
\end{figure}

We provide more results on the changes of $\lambda$ and $w$ in training.
As shown in \cref{fig:s_1_kipan_stvm_weight} ($t$ = KIPAN), four models of other cancers (HNSC, STES, CESC, and SARC) are leveraged in task vector mixup ($\lambda_i<0.8$) and aggregation ($w_i>0$).
When $t$ = LUNG (\cref{fig:s_2_lung_stvm_weight}), there are three models of other cancers (HNSC, STES, and CESC) involved more in $\mathcal{M}_t^{*}$.

\begin{table*}[htbp]
\centering
\caption{\textbf{Comparison with MIL networks for WSI-based survival prediction} on 13 cancer datasets.}
\label{tab:s_3_cmp_other_mil}
\resizebox{\textwidth}{!}{
\begin{tabular}{l|ccccccccccccc|c}
\toprule
\textbf{Method} & BRCA & KIPAN & LUNG & \makecell[c]{GBM\\ LGG} & \makecell[c]{COAD\\ READ} & STES & UCEC & HNSC & SKCM & BLCA & LIHC & CESC & SARC & \textbf{Avg.} \\
\midrule
\multirow{2}{*}{ABMIL} & 0.6648 & 0.8094 & 0.5496 & 0.7756 & 0.6725 & \underline{0.6648} & 0.7098 & 0.6201 & 0.5708 & \underline{0.6438} & 0.7265 & 0.6500 & 0.5312 & \multirow{2}{*}{0.6609}  \\ 
 & ($\pm$ 0.032) & ($\pm$ 0.013) & ($\pm$ 0.034) & ($\pm$ 0.020) & ($\pm$ 0.028) & ($\pm$ 0.040) & ($\pm$ 0.043) & ($\pm$ 0.047) & ($\pm$ 0.037) & ($\pm$ 0.015) & ($\pm$ 0.048) & ($\pm$ 0.058) & ($\pm$ 0.056) &  \\ 
\multirow{2}{*}{TransMIL} & 0.6802 & 0.8121 & 0.5669 & \textbf{0.7873} & 0.6651 & 0.6582 & \underline{0.7471} & \underline{0.6315} & \underline{0.5786} & 0.6173 & 0.7385 & 0.6365 & 0.5152 & \multirow{2}{*}{0.6642} \\
& ($\pm$ 0.026) & ($\pm$ 0.018) & ($\pm$ 0.021) & ($\pm$ 0.019) & ($\pm$ 0.059) & ($\pm$ 0.056) & ($\pm$ 0.072) & ($\pm$ 0.014) & ($\pm$ 0.039) & ($\pm$ 0.029) & ($\pm$ 0.044) & ($\pm$ 0.068) & ($\pm$ 0.042) &  \\
\multirow{2}{*}{ILRA} & 0.6645 & 0.8017 & 0.5374 & 0.7739 & 0.6147 & 0.6451 & 0.7442 & 0.5262 & 0.5586 & \textbf{0.6608} & 0.7231 & 0.6336 & 0.5351 & \multirow{2}{*}{0.6476} \\
& ($\pm$ 0.026) & ($\pm$ 0.016) & ($\pm$ 0.062) & ($\pm$ 0.016) & ($\pm$ 0.041) & ($\pm$ 0.053) & ($\pm$ 0.057) & ($\pm$ 0.024) & ($\pm$ 0.042) & ($\pm$ 0.046) & ($\pm$ 0.044) & ($\pm$ 0.016) & ($\pm$ 0.055) &  \\
\multirow{2}{*}{R$^2$T-MIL} & 0.6619 & 0.8071 & 0.5805 & 0.7775 & \underline{0.6843} & 0.6472 & 0.7253 & 0.6169 & 0.5376 & 0.6014 & 0.7163 & 0.6386 & 0.5326 & \multirow{2}{*}{0.6559} \\
& ($\pm$ 0.040) & ($\pm$ 0.021) & ($\pm$ 0.055) & ($\pm$ 0.019) & ($\pm$ 0.039) & ($\pm$ 0.033) & ($\pm$ 0.064) & ($\pm$ 0.037) & ($\pm$ 0.056) & ($\pm$ 0.024) & ($\pm$ 0.039) & ($\pm$ 0.090) & ($\pm$ 0.032) &  \\
\multirow{2}{*}{Patch-GCN} & \underline{0.7017} & \underline{0.8181} & \textbf{0.5850} & \underline{0.7853} & 0.6675 & 0.6575 & 0.7369 & 0.6019 & 0.5770 & 0.6164 & \underline{0.7487} & \underline{0.6585} & \underline{0.5736} & \multirow{2}{*}{\underline{0.6714}} \\
& ($\pm$ 0.024) & ($\pm$ 0.009) & ($\pm$ 0.027) & ($\pm$ 0.019) & ($\pm$ 0.041) & ($\pm$ 0.035) & ($\pm$ 0.058) & ($\pm$ 0.037) & ($\pm$ 0.040) & ($\pm$ 0.025) & ($\pm$ 0.062) & ($\pm$ 0.065) & ($\pm$ 0.034) &  \\  \midrule  
& \textbf{0.7408} & \textbf{0.8228} & \underline{0.5836} & 0.7763 & \textbf{0.7069} & \textbf{0.6698} & \textbf{0.7830} & \textbf{0.6569} & \textbf{0.5948} & 0.6413 & \textbf{0.7585} & \textbf{0.6965} & \textbf{0.6024} &  \\
\multirow{-2}{*}{\textbf{$\ours$}} & ($\pm$ 0.060) & ($\pm$ 0.015) & ($\pm$ 0.042) & ($\pm$ 0.022) & ($\pm$ 0.029) & ($\pm$ 0.050) & ($\pm$ 0.072) & ($\pm$ 0.042) & ($\pm$ 0.027) & ($\pm$ 0.038) & ($\pm$ 0.036) & ($\pm$ 0.065) & ($\pm$ 0.070) & \multirow{-2}{*}{\textbf{0.6949}} \\
\bottomrule
\end{tabular}
}
\end{table*}

\subsection{Task vector mixup}

\begin{figure}[H]
  \centering
   \includegraphics[width=\linewidth]{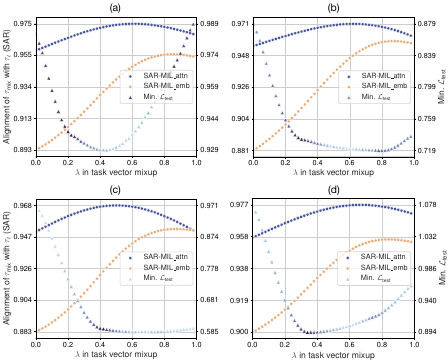}
   \caption{\textbf{Additional results of task vector mixup}.}
   \label{fig:s_3_more_sar}
\end{figure}

As presented in \cref{fig:s_3_more_sar}, similar results are observed on additional task vector pairs.
In particular, \cref{fig:s_3_more_sar}(c) shows a marginal benefit (\ie, a relatively small improvement in $\mathcal{L}_{\text{test}}$) from task vector mixup; such pairs could be filtered in the later sparse aggregating stage. 

\subsection{Hyper-parameter $K$}

As shown in \cref{tab:s_2_hyper_param_k}, using more than half of task vector mixtures often leads to a performance drop. This further confirms the effectiveness of imposing sparsity on $\tau_{\text{mix}}$.

\begin{table}[htbp]
\centering
\caption{\textbf{$\ours$ with different $K$} (the number of selected task vector mixtures in sparse aggregation).}
\label{tab:s_2_hyper_param_k}
\resizebox{\linewidth}{!}{
\begin{tabular}{lccccccc}
\toprule
$K$ & $1$ & $3$ & $5$ & $7$ & $9$ & $11$ & $13$ \\
\midrule
\textbf{Avg.} & 0.6882 & 0.6929 & \textbf{0.6949} & 0.6935 & 0.6932 & 0.6926 & 0.6912 \\
\bottomrule
\end{tabular}
}
\end{table}

\subsection{Comparison with MIL networks}

We further compare $\ours$ with representative MIL networks to see whether it is competitive with popular MIL methods. These networks are based on the Transformer or graph convolution architecture. We observed that their original settings often lead to severe overfitting (perfect training metrics), especially on those datasets with fewer than 400 cases. Moreover, they struggle to process WSIs with over 200,000 instances (see \cref{tab:s_1_data}) in mini-batch training. To address these issues, we adapt their settings to this task for stable training and apply patch sampling to ultra-large WSIs. Refer to our source code for more details.

From \cref{tab:s_3_cmp_other_mil} and \cref{tab:s_4_efficiency}, we observe that $\ours$ outperforms existing MIL networks, with an overall improvement of 3.5\% over the best baseline on 13 datasets.
Besides, Patch-GCN~\cite{chen2021whole} is a competitive network in terms of predictive performance and model efficiency. This implies that merging Patch-GCN models may lead to better performance. We leave it as our future work.

\begin{table}[htbp]
\centering
\caption{\textbf{Comparison with MIL networks in terms of model efficiency}. A WSI with 10,000 patches is used in model inference.}
\label{tab:s_4_efficiency}
\resizebox{0.85\linewidth}{!}{
\begin{tabular}{lcc}
\toprule
\multirow{2}{*}{\textbf{Method}} & \textbf{Learnable param.} & \textbf{Test-time inference} \\ 
& (M) & (GFLOPs) \\ \midrule
ABMIL & \textbf{1.20} & \underline{23.6} \\
TransMIL & 2.95 & 60.3 \\
ILRA & 4.21 & 55.3 \\
R$^2$T-MIL & 2.98 & 40.7 \\
Patch-GCN & 1.69 & \textbf{21.7} \\ \midrule
\textbf{$\ours$} & \underline{1.34}  & 40.1 \\
\bottomrule
\end{tabular}
}
\end{table}

\subsection{Complete ablation results on all datasets}

We present the complete results of key ablation components in \cref{tab:s_5_full_abl}. The contribution of key designs can be observed in most cases: hypernetwork-driven $\lambda$ on 11 out of 13 datasets and $w$ on 12 out of 13 datasets, TVM over $\lambda=1$ on 8 out of 13 datasets, and sparse aggregation on 11 out of 13 datasets.

\begin{table*}[htbp]
\centering
\caption{Complete ablation results on 13 datasets.}
\label{tab:s_5_full_abl}
\resizebox{\textwidth}{!}{
\begin{tabular}{l|ccccccccccccc|c}
\toprule
\textbf{Ablation} & BRCA & KIPAN & LUNG & \makecell[c]{GBM\\ LGG} & \makecell[c]{COAD\\ READ} & STES & UCEC & HNSC & SKCM & BLCA & LIHC & CESC & SARC & \textbf{Avg.} \\
\midrule
w/o mixup & 0.7183 & 0.8264 & 0.5876 & 0.7745 & 0.6971 & 0.6781 & 0.7840 & 0.6420 & 0.5954 & 0.6387 & 0.7400 & 0.6886 & 0.5358 & \multirow{2}{*}{0.6851}  \\ 
($\lambda=1$) & ($\pm$0.051) & ($\pm$0.016) & ($\pm$0.039) & ($\pm$0.024) & ($\pm$0.024) & ($\pm$0.030) & ($\pm$0.069) & ($\pm$0.022) & ($\pm$0.037) & ($\pm$0.038) & ($\pm$0.046) & ($\pm$0.043) & ($\pm$0.070) &  \\ \midrule 
w/o $\mathcal{H}_{\text{mix}}$ & 0.7323 & 0.8152 & 0.5901 & 0.7728 & 0.7044 & 0.6645 & 0.7852 & 0.6548 & 0.5903 & 0.6404 & 0.7579 & 0.6951 & 0.5944 & \multirow{2}{*}{0.6921} \\
(param. $\lambda$) & ($\pm$0.061) & ($\pm$0.019) & ($\pm$0.040) & ($\pm$0.024) & ($\pm$0.027) & ($\pm$0.047) & ($\pm$0.073) & ($\pm$0.041) & ($\pm$0.031) & ($\pm$0.038) & ($\pm$0.034) & ($\pm$0.066) & ($\pm$0.066) &  \\ \midrule
\multirow{2}{*}{w/o sparse} & 0.7263 & 0.8212 & 0.5865 & 0.7746 & 0.7009 & 0.6640 & 0.7803 & 0.6526 & 0.5891 & 0.6433 & 0.7541 & 0.6937 & 0.5994 & \multirow{2}{*}{0.6912} \\
&($\pm$0.052) &($\pm$0.016) &($\pm$0.043) &($\pm$0.021) &($\pm$0.027) &($\pm$0.055) &($\pm$0.075) &($\pm$0.038) &($\pm$0.024) &($\pm$0.041) &($\pm$0.034) &($\pm$0.063) &($\pm$0.086) &  \\ \midrule
w/o $\mathcal{H}_{\text{agg}}$ & 0.7133 & 0.8042 & 0.5417 & 0.7800 & 0.6564 & 0.5984 & 0.7359 & 0.5266 & 0.5753 & 0.6289 & 0.7035 & 0.6424 & 0.5299 & \multirow{2}{*}{0.6490} \\
(param. $w$) & ($\pm$0.015) & ($\pm$ 0.014) & ($\pm$0.033) & ($\pm$0.017) & ($\pm$0.033) & ($\pm$0.036) & ($\pm$0.049) & ($\pm$0.037) & ($\pm$0.032) & ($\pm$0.049) & ($\pm$0.034) & ($\pm$0.025) & ($\pm$0.030) &  \\ \midrule
& 0.7408 & 0.8228 & 0.5836 & 0.7763 & 0.7069 & 0.6698 & 0.7830 & 0.6569 & 0.5948 & 0.6413 & 0.7585 & 0.6965 & 0.6024 &  \\
\multirow{-2}{*}{\textbf{$\ours$}} & ($\pm$0.060) & ($\pm$0.015) & ($\pm$0.042) & ($\pm$0.022) & ($\pm$0.029) & ($\pm$0.050) & ($\pm$0.072) & ($\pm$0.042) & ($\pm$0.027) & ($\pm$0.038) & ($\pm$0.036) & ($\pm$0.065) & ($\pm$0.070) & \multirow{-2}{*}{\textbf{0.6949}} \\
\bottomrule
\end{tabular}
}
\end{table*}

\begin{figure*}[htbp]
\centering
\includegraphics[width=0.92\textwidth]{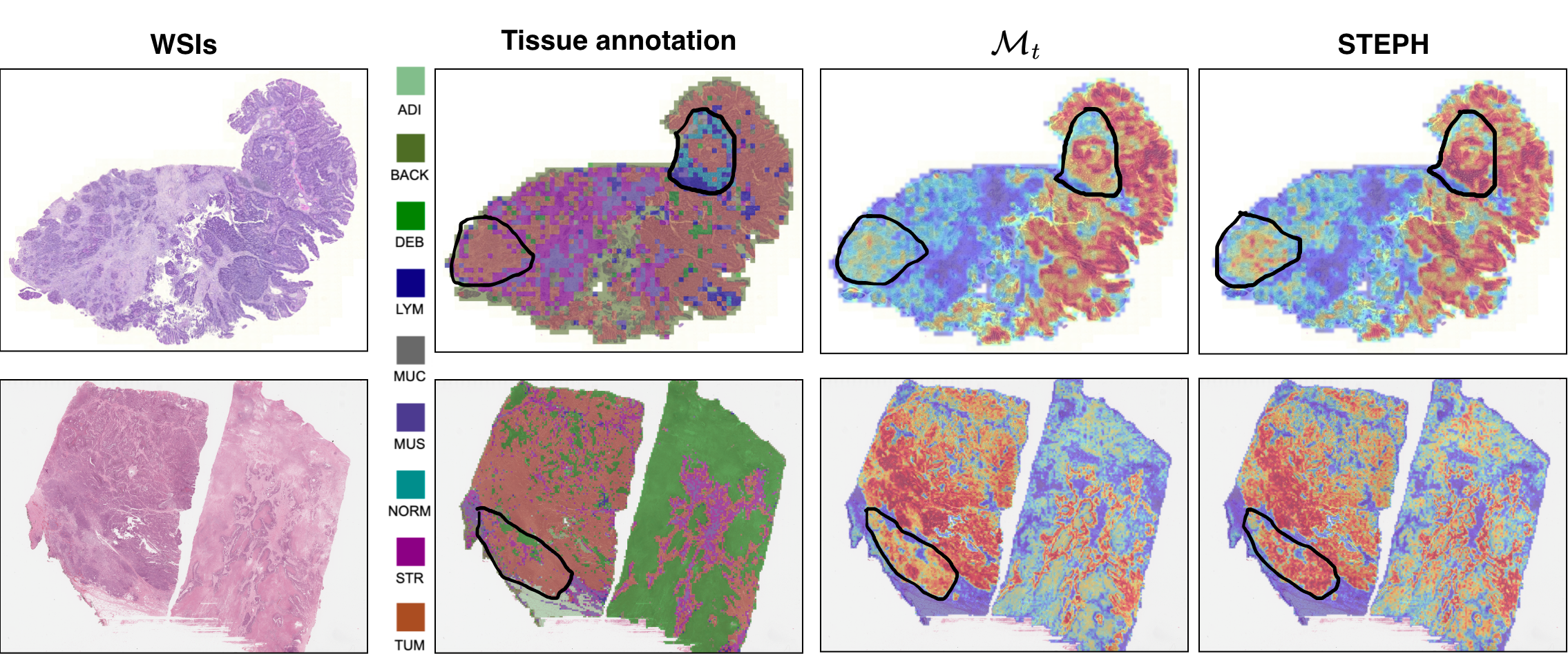}
\caption{Visualization of attention maps given by models. The two WSIs are from the test set of COADREAD. Tumor (TUM).}
\label{fig:s_4_vis}
\end{figure*}

\subsection{Training efficiency}

\textbf{Comparison with other model merging methods}~To show the training efficiency of $\ours$, we measure the mean training time per epoch on BRCA and report it for all model merging methods.
\cref{tab:s_6_train_time} shows that $\ours$ only introduces marginal training costs compared to the methods with AdaMerging.

\begin{table}[htbp]
\centering
\tabcolsep=0.1cm
\caption{Mean training time per epoch on BRCA.}
\label{tab:s_6_train_time}
\resizebox{\linewidth}{!}{
\begin{tabular}{lcccccc}
\toprule
\textbf{Method} & Model Avg. & AdaMerging & TIES AM & Surgery AM & Iso-C AM & STEPH \\
\midrule
\multirow{2}{*}{\textbf{Time} (s)}  & 0.00 & 78.08 & 78.42 & 78.00 & 78.50 & \textbf{80.45} \\
& ($\pm$ 0.00) & ($\pm$ 0.92) & ($\pm$ 2.20) & ($\pm$ 1.34) & ($\pm$ 1.54) & ($\pm$ 1.63) \\
\bottomrule
\end{tabular}
}
\end{table}

\noindent\textbf{Increasing the number of source models}~We further vary the number of source models (denoted by $m$) to show the training efficiency of $\ours$. The number of chosen task vector mixtures, $K$, is set to be equal to $m$, \textit{i.e.}, all source models are utilized.
\cref{tab:s_7_train_time} shows that increasing $m$ would not introduce too much training time. 

\begin{table}[htbp]
\centering
\tabcolsep=0.1cm
\caption{Mean training time per epoch on BRCA with different numbers of source models ($m$).}
\label{tab:s_7_train_time}
\resizebox{\linewidth}{!}{
\begin{tabular}{lcccccccc}
\toprule
$m$ ($K$) & 6 & 7 & 8 & 9 & 10 & 11 & 12 \\
\midrule
\multirow{2}{*}{\textbf{Time} (s)} & 72.91 & 72.98 & 74.51 & 75.62 & 75.60 & 75.74 & \textbf{76.69} \\
& ($\pm$ 1.20) & ($\pm$ 1.42) & ($\pm$ 3.00) & ($\pm$ 1.54) & ($\pm$ 1.62) & ($\pm$ 2.18) & ($\pm$ 1.79) \\
\bottomrule
\end{tabular}
}
\end{table}

\subsection{Interpretability}

We visualize and compare the attention maps given by $\mathcal{M}_t$ and $\ours$, following the setting of \cite{liu2025cross}. From \cref{fig:s_4_vis}, we observe that $\ours$ captures more prognostic cues: tumor areas and the boundary of tumor infiltration.

\section{More discussions}

\begin{figure}[tp]
  \centering
  \includegraphics[width=\linewidth]{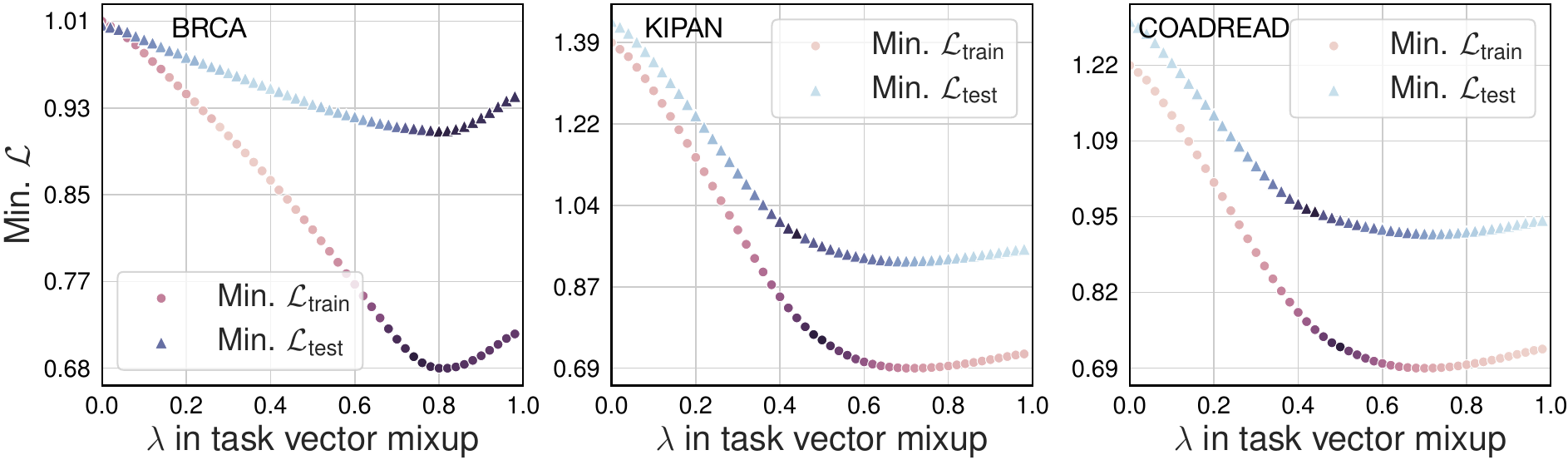}
   \caption{Loss of TVM between BLCA and others (w/ top $w$).}
   \label{fig:s_5_blca_tvm}
\end{figure}

\textbf{Comparison with ROUPKT}~\cite{liu2025cross}~$\ours$ is based on model merging, whereas ROUPKT is based on representation transfer (as shown in \cref{fig:1_existing_frameworks}). Two methods are distinct in design philosophy, though both are used for knowledge transfer.
In the comparative results given by \cref{tab:1_main_cmp_results}, ROUPKT performs better than $\ours$ on 4 datasets.
We examine the cause of the largest performance gap, BLCA, by analyzing the loss landscape of TVM. From \cref{fig:s_5_blca_tvm}, we observe that $\lambda$ with the best $\mathcal{L}_{\text{test}}$ is often larger (0.7$\sim$0.8). It would be helpful to decrease the degree of penalty by setting a smaller $\beta$ for $\mathcal{L}_{\text{mix}}$ to encourage a larger $\lambda$. However, $\beta$ is simply set to 0.05 for all datasets in this paper without fine-tuning. Fine-tuning $\beta$ on BLCA may improve the performance of $\ours$.

\noindent\textbf{Choice of foundation models}~UNI2-h~\cite{chen2024towards} is adopted because \textit{i}) it is trained with private data (not TCGA) so data leakage can be avoided and \textit{ii}) it has demonstrated excellent generalizability. Other models, like CONCH~\cite{lu2024visual}, are also competitive. Yet, one more foundation model leads to substantial computational costs since there are 8,818 WSIs ($>$100 million patches) and 5-fold cross-validation experiments on 13 datasets.
Given these, we choose one of the representatives.

\section{Reproducibility statement}

To ensure reproducibility, we provide detailed descriptions of the datasets in \cref{sec:c_1} and comprehensive implementation specifics in \cref{sec:c_2}. Our source code is publicly available at \url{https://github.com/liupei101/STEPH}.
It includes the checkpoints and training logs of our model.

\end{appendix}

\end{document}